\documentclass[preprint,12pt,authoryear]{elsarticle}

\usepackage{amsfonts}
\usepackage[ruled,vlined]{algorithm2e}

\usepackage{array}
\usepackage[caption=false,font=normalsize,labelfont=sf,textfont=sf]{subfig}
\usepackage{textcomp}
\usepackage{stfloats}
\usepackage{url}
\usepackage{verbatim}
\usepackage{graphicx}

\usepackage{amsmath}
\usepackage{amssymb}
\usepackage{booktabs}
\usepackage{multirow}
\usepackage[normalem]{ulem}
\usepackage{chngcntr}
\usepackage{xcolor}
\usepackage{pifont}
\usepackage{tabularx}
\usepackage{geometry}

\usepackage{makecell}
\usepackage{adjustbox}

\usepackage{hyperref}

\useunder{\uline}{\ul}{}
\DeclareMathOperator*{\argmax}{arg\,max}  

\usepackage[capitalize]{cleveref}
\crefname{section}{Section}{Sections}
\crefname{table}{Table}{Tables}
\crefname{figure}{Figure}{Figures}
\crefname{equation}{Eq.}{Eqs.}
\crefname{appendix}{Appendix.}{Appendix.}

\journal{Engineering Applications of Artificial Intelligence}

\begin{document}

\begin{frontmatter}



\title{Avoid Wasted Annotation Costs in Open-set Active Learning with Pre-trained Vision-Language Model}

\author[label1]{Jaehyuk~Heo}
\ead{jaehyuk.heo@snu.ac.kr}
\author[label1]{Pilsung~Kang\corref{cor1}}
\ead{pilsung\_kang@snu.ac.kr}
\affiliation[label1]{organization={Department of Industrial Engineering, Seoul National University},
            country={Republic of Korea}}

\cortext[cor1]{Corresponding author}

\begin{abstract}
Active learning (AL) aims to enhance model performance by selectively collecting highly informative data, thereby minimizing annotation costs. However, in practical scenarios, unlabeled data may contain out-of-distribution (OOD) samples, which are not used for training, leading to wasted annotation costs if data is incorrectly selected. Therefore, to make active learning feasible in real-world applications, it is crucial to consider not only the informativeness of unlabeled samples but also their purity to determine whether they belong to the in-distribution (ID). Recent studies have applied AL under these assumptions, but challenges remain due to the trade-off between informativeness and purity, as well as the heavy dependence on OOD samples. These issues lead to the collection of OOD samples, resulting in a significant waste of annotation costs. To address these challenges, we propose a novel query strategy, VLPure-AL, which minimizes cost losses while reducing dependence on OOD samples. VLPure-AL sequentially evaluates the purity and informativeness of data. First, it utilizes a pre-trained vision-language model to detect and exclude OOD data with high accuracy by leveraging linguistic and visual information of ID data. Second, it selects highly informative data from the remaining ID data, and then the selected samples are annotated by human experts. Experimental results on datasets with various open-set conditions demonstrate that VLPure-AL achieves the lowest cost loss and highest performance across all scenarios.
\end{abstract}




\end{frontmatter}



\section{Introduction}
\label{sec:intro}

Active Learning (AL) is a research area proposed to improve model performance while minimizing the cost of data acquisition required for model training~\citep{al_survey}. Although deep learning models have achieved impressive results on various tasks such as classification and regression, their success often depends on significant annotation costs~\citep{deep_learning, imagenet, laion}. To address this issue, AL introduces diverse query strategies, designed to identify the most informative data samples from large pools of unlabeled data, thereby enhancing performance with minimal user cost~\citep{conf, ll, coreset, intro3}. AL has been extensively investigated and applied across a variety of fields such as action recognition~\citep{intro1} and visual tracking~\citep{intro2}.

\begin{figure}[t!] 
\begin{center}
\includegraphics[width=1\linewidth]{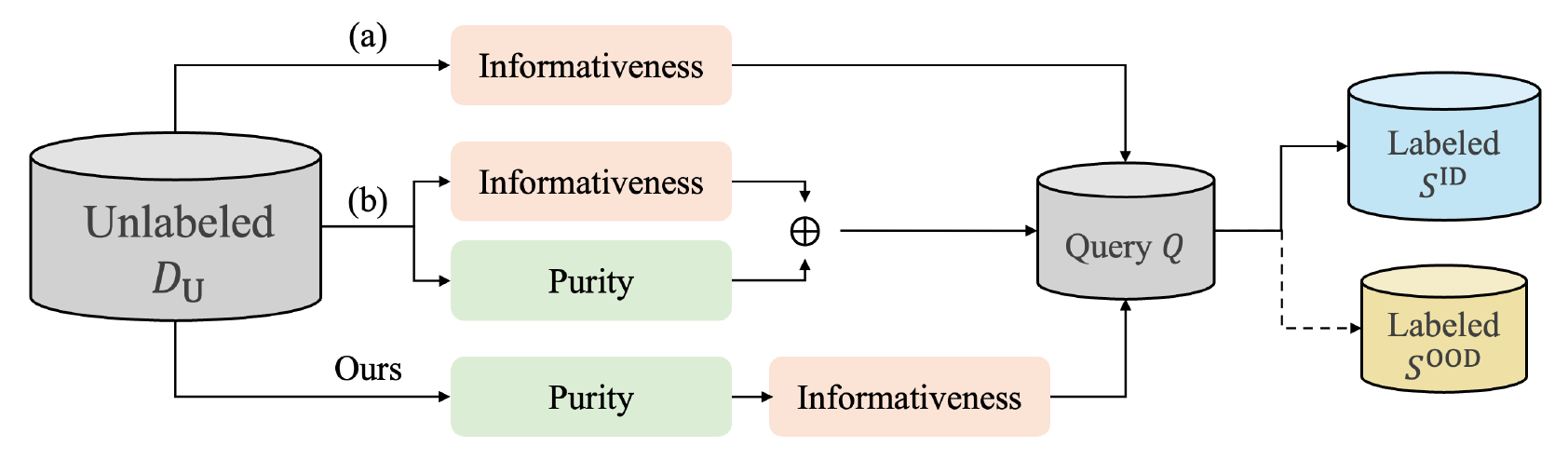}
\end{center}
\vspace{-20pt}
\caption{Comparison of data selection criterion for different active learning scenarios: (a) uses informativeness as the selection criteria for standard AL, while (b) and Ours use both informativeness and purity as the selection criteria for open-set AL. Unlike (b), Ours evaluates purity and informativeness sequentially.}
\label{fig:al_method}
\end{figure}

\begin{figure}[t!]
\begin{center}
\includegraphics[width=1\linewidth]{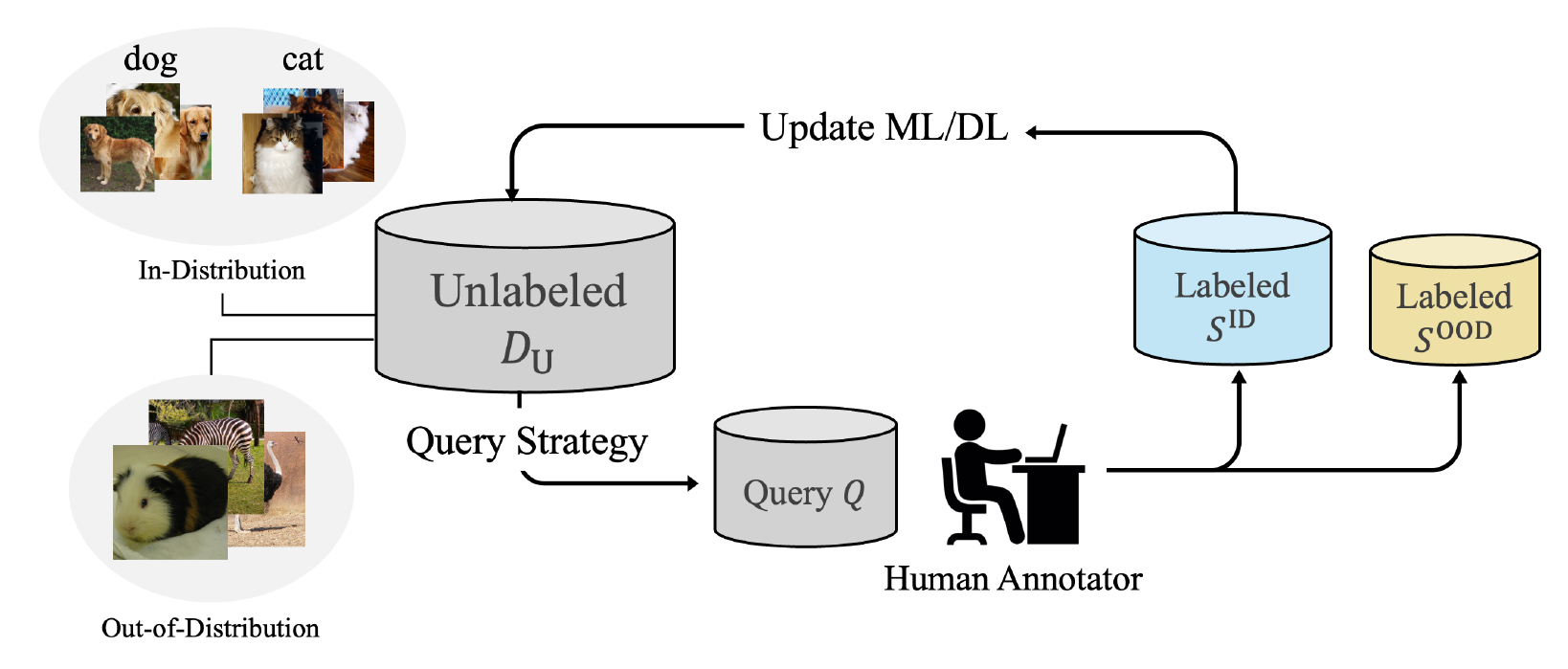}
\end{center}
\vspace{-20pt}
\caption{Open-set AL process.}
\label{fig:al_process}
\end{figure}

Traditional AL research, termed \textit{standard AL}~\citep{conf, ll, coreset, probcover}, assumes that the unlabeled data is collected from the same domain as the training data called \textit{in-distribution data (ID)}. As shown in \cref{fig:al_method}~(a), the query strategy of standard AL selects an unlabeled sample focused on its \textit{informativeness}, which refers to the degree to which the sample can help improve model performance by providing information that the model has not yet learned. However, it is difficult to assume that unlabeled data only contains ID samples. For instance, if the classification task involves distinguishing between dogs and cats, it is assumed that the unlabeled data only contains images of dogs and cats. However, diverse data collected online might include other animals like guinea pigs or zebras (\cref{fig:al_process}). If the unlabeled data includes out-of-distribution (OOD) samples and only \textit{informativeness} is considered, the selected data might include OOD samples, which cannot be used for training, leading to wasted annotation costs. Therefore, to practically implement active learning, it is essential to consider the \textit{purity} of the selected data from the unlabeled pool, ensuring that it shares the same distribution as the training data. In other words, it is important to minimize wasted annotation costs to ensure the collection of more in-distribution samples, which is essential for improving model performance.

Recently, AL research has expanded to focus on the \textit{purity} of unlabeled samples under the assumption that the unlabeled data pool contains not only ID samples but also OOD samples, as illustrated in \cref{fig:al_method}~(b), and it is called open-set AL~\citep{ccal, mqnet, lfosa, eoal}. Previous studies on open-set AL proposed query strategies that balance \textit{purity} and \textit{informativeness} through contrastive learning and meta-learning, adjusting the importance of these two metrics depending on the data collection stage to address their inherent dilemma~\citep{ccal, mqnet}. However, these methods face limitations, as they cannot simultaneously satisfy both \textit{purity} and \textit{informativeness} at every stage, resulting in an increasing proportion of OOD samples in the collected data. Another approach focuses on estimating the \textit{purity} of unlabeled samples by training an OOD detector while simultaneously incorporating \textit{informativeness} into the query strategy~\citep{lfosa, eoal}. This approach also has limitations, as it requires a significant number of OOD samples to improve the performance of the OOD detector. Both approaches share a common drawback in that they require or heavily depend on collecting unnecessary data to satisfy the selection criteria for training samples.

We propose a sequential approach framework to address the aforementioned issues by assessing \textit{purity} and then applying \textit{informativeness} as shown in \cref{fig:al_method}. This sequential approach alleviates the trade-off problem and allows the flexible utilization of various query strategies proposed in standard AL while considering \textit{purity}. However, achieving high \textit{purity} accuracy is essential to minimize cost loss in the sequential process. To achieve high \textit{purity} accuracy, we propose \textbf{VLPure-AL}, which leverages Contrastive Language-Image Pre-training (CLIP)~\citep{clip} while reducing dependence on OOD samples.

CLIP is trained to align representations of images and text in the same embedding space, enabling the classification of images using textual information. Due to these advantages, recent research has actively employed CLIP in the OOD detection domain~\citep{zoc, mcm, clipn}.  In VLPure-AL, we enhance the process of assessing the \textit{purity} of unlabeled samples by utilizing the textual information of CLIP's text encoder and applying visual similarity weights, which leverage class-specific visual information from the collected labeled data. Furthermore, for more accurate \textit{purity} assessment, we refine the method by incorporating unavoidable OOD samples through self-temperature tuning, which optimizes the temperature scaling factor used in the \textit{purity} assessment process.

The advantages of our proposed method can be summarized as follows:

\begin{itemize}
    \item By sequentially evaluating purity with high accuracy and considering informativeness, it resolves the trade-off issue between purity and informativeness while allowing the flexible application of Standard AL methods.
    \item VLPure-AL reduces dependence on OOD samples by leveraging a pre-trained CLIP model and minimizes cost loss through accurate OOD detection using textual information, visual similarity weights, and self-temperature tuning.
    \item The proposed method demonstrates lower annotation cost loss and superior performance compared to both standard AL and open-set AL methods under various open-set conditions.
\end{itemize}

This paper is organized as follows: \cref{sec:rel_works} provides a review of standard AL and open-set AL. \cref{sec:pro_method} demonstrates the workflow of the proposed framework and introduces our proposed query strategy for open-set AL. \cref{sec:exp} reports experimental results under various open-set conditions. Finally, \cref{sec:conclusion} summarizes our contributions and discusses the limitations and future direction of our work.

\begin{figure}[t!] 
\begin{center}
\includegraphics[width=1\linewidth]{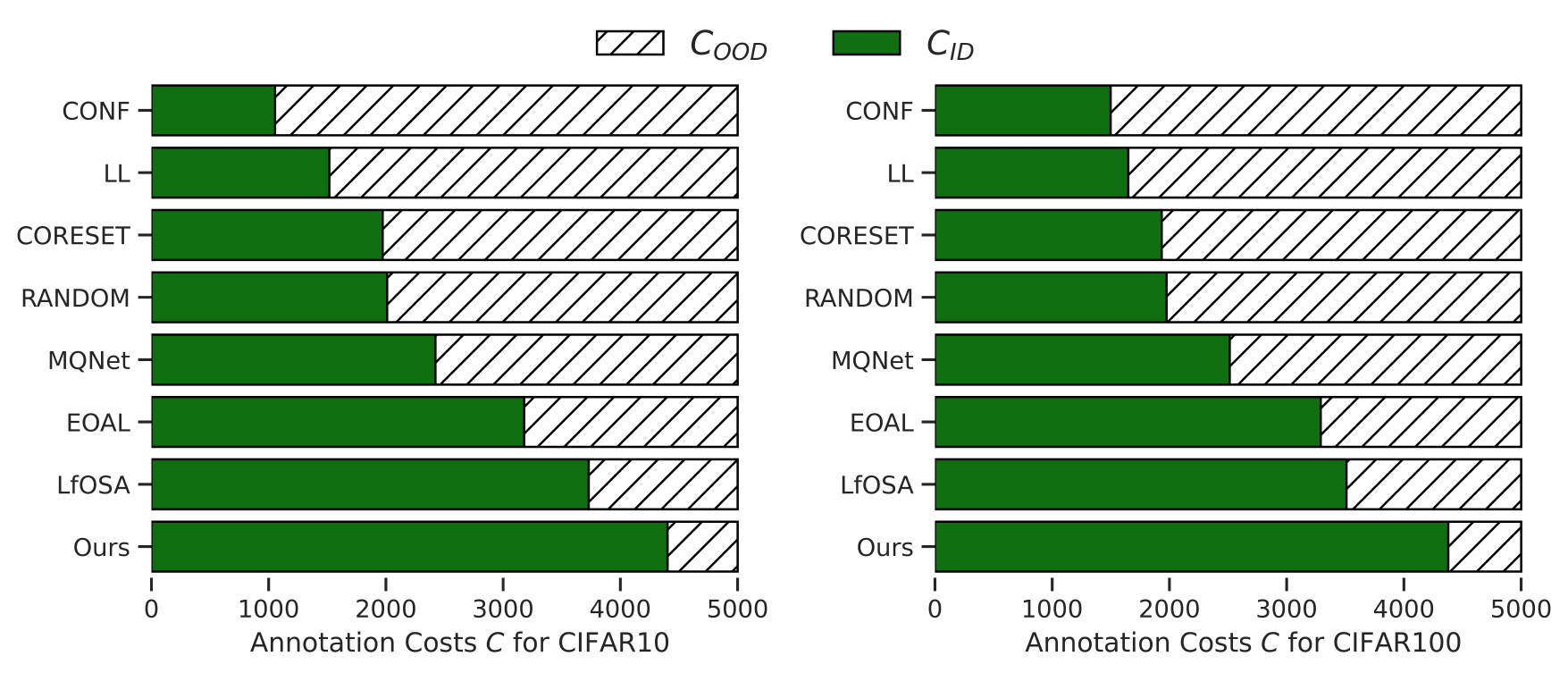}
\end{center}
\vspace{-20pt}
\caption{Comparison of wasted annotation costs in relation to total annotation costs incurred when applying various AL methods to CIFAR10 and CIFAR100 configured as open-set data.
$C_\text{OOD}$ represents the cost incurred when selected data is OOD data, and $C_\text{ID}$ represents the cost for ID data. Standard AL methods such as CONF, LL, and CORESET result in higher cost losses compared to RANDOM due to the higher selection of OOD data. On the other hand, existing open-set AL methods like MQNet, EOAL, and LfOSA show less cost loss than RANDOM but still suffer significant losses relative to the total annotation costs.}
\label{fig:costs}
\end{figure}

\section{Related Works}
\label{sec:rel_works}

\subsection{Standard Active Learning} \label{sec:rel_standard_al}
Standard AL aims to reduce annotation costs by selecting only the most informative data from a large pool of unlabeled data to enhance model performance. The query strategies proposed to select data are primarily based on the informativeness of the data.

The criteria for determining informativeness in query strategies can be broadly divided into two classes. The first is \textit{uncertainty-based query strategy}~\citep{conf, ll}. In a classification task, the uncertainty-based strategy assesses the degree of confidence in classifying unlabeled data based on the decision boundary of the trained model. If the model's certainty about the data is low, it implies that the model finds it difficult to make an accurate judgment. Therefore, data with high uncertainty is selected for model training. The second is \textit{diversity-based query strategy}~\citep{coreset, probcover}. Despite the large data size, much of it contains redundant information. Hence, a small subset of data selected based on diversity can provide information similar to that of a larger dataset. Therefore, strategies that consider diversity select data that differ from the current data based on a comparison with labeled data. 

However, if the unlabeled data contains OOD data, both criteria can lead to selecting data that cannot be used for training, causing cost losses. It is difficult to determine ID data based solely on high uncertainty. This is because OOD samples, which were not seen during model training, also have high uncertainty. OOD samples are selected based solely on the diversity of representation because the representation of OOD samples is located in a different space compared to ID samples.

\subsection{Open-set Active Learning} \label{sec:rel_openset_al}

Unlike the assumptions in standard AL, open-set AL considers the presence of OOD data in the unlabeled pool. Various methods have been proposed for this purpose. Firstly, CCAL~\citep{ccal} and MQNet~\citep{mqnet} perform contrastive learning on unlabeled data to determine data purity using the trained model. CCAL uses SimCLR~\citep{simclr} and CSI~\citep{csi} to reflect purity and informativeness, respectively, and proposes a joint query strategy that combines these scores for selecting data. However, contrastive learning alone focuses on purity, making it difficult to select highly informative data. To resolve this trade-off, MQNet adjusts the purity obtained from CSI and the informativeness from standard AL using meta-learning, allowing data collection to prioritize purity or informativeness as needed.

Secondly, research has proposed training OOD detection models to assess data purity. LfOSA~\citep{lfosa} trains a classifier to distinguish $K+1$ classes, including $K$ ID classes and one additional OOD class, using labeled ID and OOD data. The classifier predicts the class for each sample of unlabeled data, and based on the maximum activation value of the predicted class, a Gaussian Mixture Model (GMM) is trained with $K$ components. Each GMM assumes two Gaussian distributions. The distribution with the higher mean is assumed to represent the ID data distribution. Therefore, the higher the probability of a sample belonging to the Gaussian distribution with the higher mean, the higher its purity. Conversely, a lower probability indicates that the sample does not belong to any ID classes, meaning it is OOD class. However, LfOSA does not consider informativeness, limiting model performance. EOAL~\citep{eoal} addresses this issue by proposing a method to select data using a combined score based on two entropy measures representing purity and informativeness. Firstly, to assess informativeness, $K$ binary classifiers are trained, and the entropy scores for the unlabeled data are calculated using these $K$ binary classifiers corresponding to the ID classes. Secondly, to assess purity, a classifier is trained to classify $K+1$ classes, and the representations from this trained classifier for the labeled OOD data are used to perform clustering. The probability of belonging to each cluster is calculated based on the distances between clusters, and the entropy for data purity is computed for the unlabeled data. Combining these two scores, EOAL selects data with high purity and informativeness.

While previous open-set AL methods show improved performance over standard AL by considering OOD data, they still incur annotation cost losses. MQNet, for instance, tends to select a large amount of OOD data because it prioritizes informativeness over purity when selecting high-informativeness data. Methods like LfOSA and EOAL require additional OOD data for OOD detector training, inherently leading to cost losses. By leveraging a pre-trained CLIP, we reduce dependence on OOD samples, accurately assess purity, and consider informativeness, effectively addressing the limitations of previous studies.

\section{Proposed Method}
\label{sec:pro_method}

\begin{figure*}[t!] 
\begin{center}
\includegraphics[width=1.\linewidth]{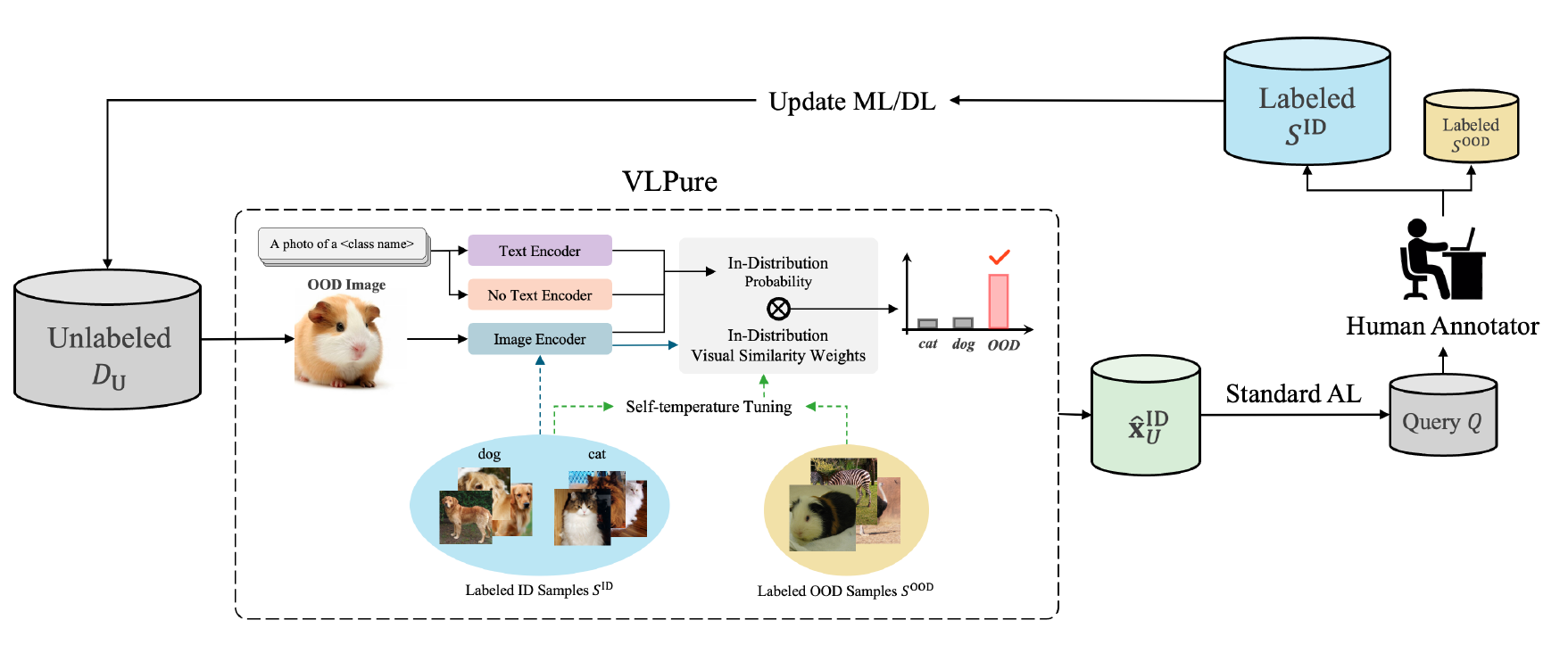}
\end{center}
\vspace{-20pt}
\caption{Overview of evaluating the purity of an image in VLPure-AL. We utilize the CLIPN to perform zero-shot OOD detection. According to CLIPN, two text encoders and an image encoder compute an image's in-distribution (ID) probability. We additionally use visual similarity weights to update ID probability using labeled ID samples $S^{\text{ID}}$. Self-temperature tuning is used to find the optimal temperature parameter for improving zero-shot OOD performance.}
\label{fig:ours_overview}
\end{figure*}

\begin{figure*}[t!] 
\begin{center}
\includegraphics[width=1\linewidth]{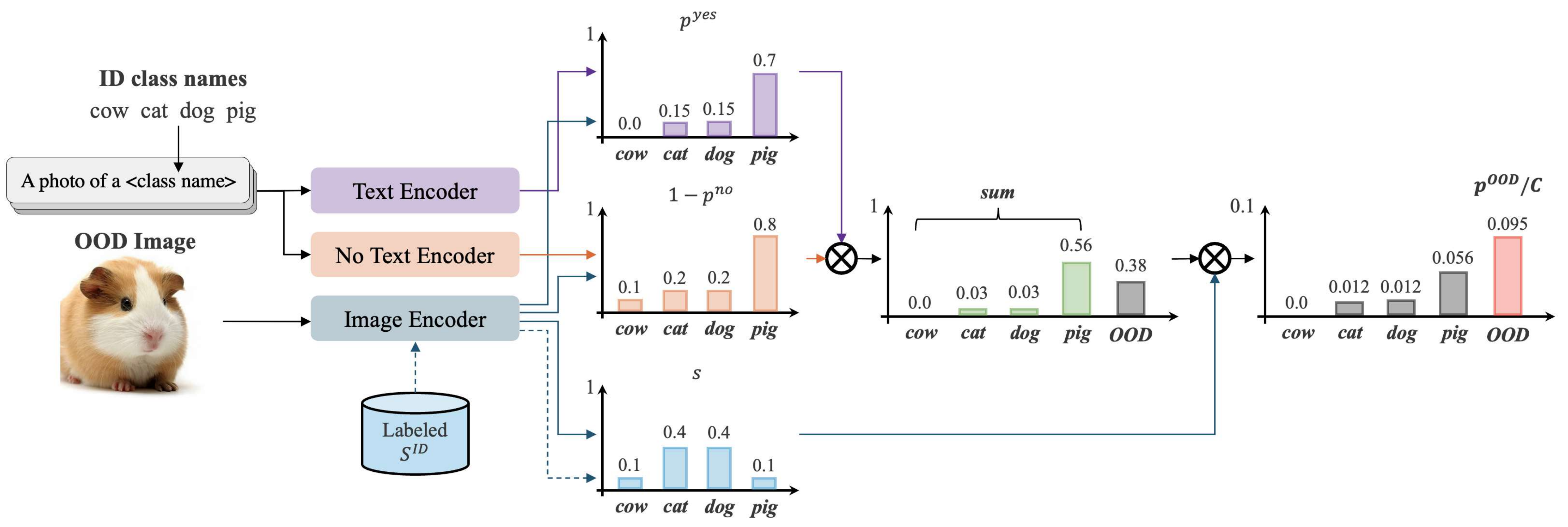}
\end{center}
\vspace{-20pt}
\caption{Example of the data purity assessment process in VLPure-AL.}
\label{fig:ours_example}
\end{figure*}

The proposed VLPure-AL comprises two sequential processes: (1) measuring data purity based on a pre-trained CLIP and (2) computing informativeness of query strategies used in standard AL. In this section, we describe how our CLIP-based method assesses the purity of unlabeled data. As illustrated in \cref{fig:ours_overview}, we use ID class names as linguistic information to calculate the ID probability of unlabeled images. Furthermore, we integrate visual information from labeled ID data to apply weighting to the ID probability. An example of VLPure can be seen in \cref{fig:ours_example}.

\subsection{Problem Statement} \label{sec:pro_problem}

The open-set AL problem defined in this study aims to classify images into $K$ classes, to improve the performance of an image classifier $f^{\text{clf}}$ based on labeled data. We define the unlabeled data targeted for collection as open-set data $D_\text{U}=\{x_i \in \textbf{x}^{\text{ID}} \cup \textbf{x}^{\text{OOD}} \}_{i=1}^{N_\text{U}}$, assuming the presence of both ID and OOD data. In AL, a fixed budget $C$ is used for annotation costs in each round of data collection, repeated $R$ times. If annotating a single image costs one unit, setting $C$ to 10 means 10 images can be annotated per round. The selected queries from $D_{\text{U}}$ are denoted as $Q$. These queries are annotated by human experts, adding ID data $\textbf{x}^{\text{ID}}$ to $S^{\text{ID}}$ and OOD data to $S^{\text{OOD}}$. The labeled ID data $S^{\text{ID}}$ is used to train the classifier $f^{\text{clf}}$. Thus, open-set AL aims to maximize model performance by minimizing the proportion of OOD data and selecting highly informative data with minimal cost.

\subsection{CLIP-based OOD Detection} \label{sec:pro_ID_prob}

Recently, large-scale datasets like MS COCO~\citep{mscoco} and LAION-2B~\citep{laion} have enabled CLIP~\citep{clip} to be used not only for ID classification but also for OOD detection~\citep{zoc, mcm, clipn}. CLIPN~\citep{clipn}, the latest method, enhances CLIP by adding a text encoder for the concept of \textit{“no”}. Unlike previous CLIP-based zero-shot OOD detection methods~\citep{zoc, mcm}, which only consider proximity to class-specific text, CLIPN incorporates the notion of dissimilarity to achieve high performance in OOD detection. Therefore, our proposed method uses CLIPN to detect OOD data in the unlabeled pool.

CLIPN assesses whether an image is OOD by considering the similarity between the image and class-specific textual information and the probability that the image does not belong to the class. First, the visual representation $f_i$ of an image $x_i$ is obtained by inputting it into the image encoder $\phi$. Next, pre-defined textual templates $t_{k,j}$ such as "A photo of $\{\text{class}_j\}$" are input into the text encoder $\psi$ for each target class to calculate the average textual representation $g_j$. Similarly, the \textit{“no”} text encoder in CLIPN calculates the average representation $g_j^{no}$ for each class.

\begin{equation} \label{eq:get_features}
\begin{aligned}
f_i &= \phi^{\text{img}}(x_i) \\
g_j &= \frac{1}{T} \sum_{l=1}^T \psi(t_{l,j}) \\
g_j^{\text{\text{no}}} &= \frac{1}{T} \sum_{l=1}^T \psi^{\text{no}}(t_{l,j}).
\end{aligned}    
\end{equation}

Next, the probability that $x_i$ belongs to each class is calculated by computing the cosine similarity between the visual and textual representations and applying softmax.

\begin{align} \label{eq:yes_prob}
p_{ij} = \frac{\exp{(f_i \cdot g_j / \tau)}}{\sum_{k=1}^{K} \exp{(f_i \cdot g_k / \tau)}},
\end{align}

\noindent where $\tau$ is a temperature parameter that adjusts model confidence. The probability that $x_i$ does not belong to each class $p_{ij}$ is defined as follows.

\begin{align} \label{eq:no_prob}
p_{ij}^{\text{no}} = \frac{\exp{(f_i \cdot g_j^{\text{no}} / \tau)}}{\exp{(f_i \cdot g_j / \tau)} + \exp{(f_i \cdot g_j^{\text{no}} / \tau)}}.
\end{align}

Finally, the probabilities $p_{ij}^{\text{ID}}$ for ID data and $p_i^{\text{OOD}}$ for OOD data are calculated using the agreeing-to-differ (ATD) method proposed in CLIPN. ATD defines $p_i^{\text{OOD}}$ based on $p_{ij}^{\text{ID}}$ without requiring a user-defined threshold.

\begin{equation} \label{eq:calc_prob}
\begin{aligned}
p_{ij}^{\text{ID}} &= p_{ij} (1 - p_{ij}^{\text{no}}) \\ 
p_i^{\text{OOD}} &= 1 - \sum_j^{K}{p_{ij} (1 - p_{ij}^{\text{no}})} \\
\mathbb{I}(x_i) &= 
\begin{cases} 
1, & p_i^{\text{OOD}} > \max_{j}{p_{ij}^{\text{ID}}}, \\
0, & \text{else},
\end{cases}
\end{aligned}
\end{equation}

\noindent where $\mathbb{I}(x_i)$ indicates whether the image $x_i^{\text{img}}$ is OOD. If $p_i^{\text{OOD}}$ is greater than the maximum $p_{ij}^{\text{ID}}$, then $\mathbb{I}(x_i)$ is 1, indicating that $x_i$ is considered OOD.

\subsection{Visual Similarity Weighting} \label{sec:pro_sim}

Although CLIPN shows high performance in detecting OOD data using textual information, it encounters challenges when the number of classes increases or when the OOD data domain is similar to that of ID data. We use visual information from ID data as additional weights to address these challenges. First, the average representation $\mu^{\text{ID}}_j$ is calculated by applying the image encoder $\phi$ to the labeled ID samples $\text{x}^{\text{ID}}_j$ for each class~$j$.

\begin{align} \label{eq:vis_lb_vec}
\mu^{\text{ID}}_{j} = \frac{1}{N_{j}^{\text{ID}}} \sum_{i=1}^{N_{j}^{\text{ID}}}{\phi(x_i^{\text{ID}})},
\end{align}

\noindent where $N_j^{\text{ID}}$ is the number of labeled samples for class $j$.

Next, the visual similarity weight $s_{ij}$ for each class is calculated by computing the cosine similarity between the visual representation $\phi$ and the average representation $\mu^{\text{ID}}_j$, and converting it to probabilities using the softmax function.

\begin{align} \label{eq:vis_weights}
s_{ij} = \frac{\exp{(f_i \cdot \mu^{\text{ID}}_j / \tau)}}{\sum_{k=1}^{K} \exp{(f_i \cdot \mu^{\text{ID}}_k / \tau)}}.
\end{align}

The visual similarity weight $s_{ij}$ is used to adjust the probabilities of ID data. The adjusted probabilities $p_i^{*, \text{ID}}$ and $p_i^{*, \text{OOD}}$ are calculated by applying the weights as follows.

\begin{equation}\label{eq:update_probs}
\begin{aligned}
p_{ij}^{*, \text{ID}} &= p_{ij}^{\text{ID}} \cdot s_{ij} \\
p_{i}^{*, \text{OOD}} &= p_{i}^{\text{OOD}} / K \\
\mathbb{I}(x_i) &= 
\begin{cases} 
1, & p_i^{*, \text{OOD}} \geq \max_{j}{p_{ij}^{*, ID}}, \\
0, & \text{else},
\end{cases}
\end{aligned}    
\end{equation}

\begin{figure}
\begin{center}
\includegraphics[width=1\linewidth]{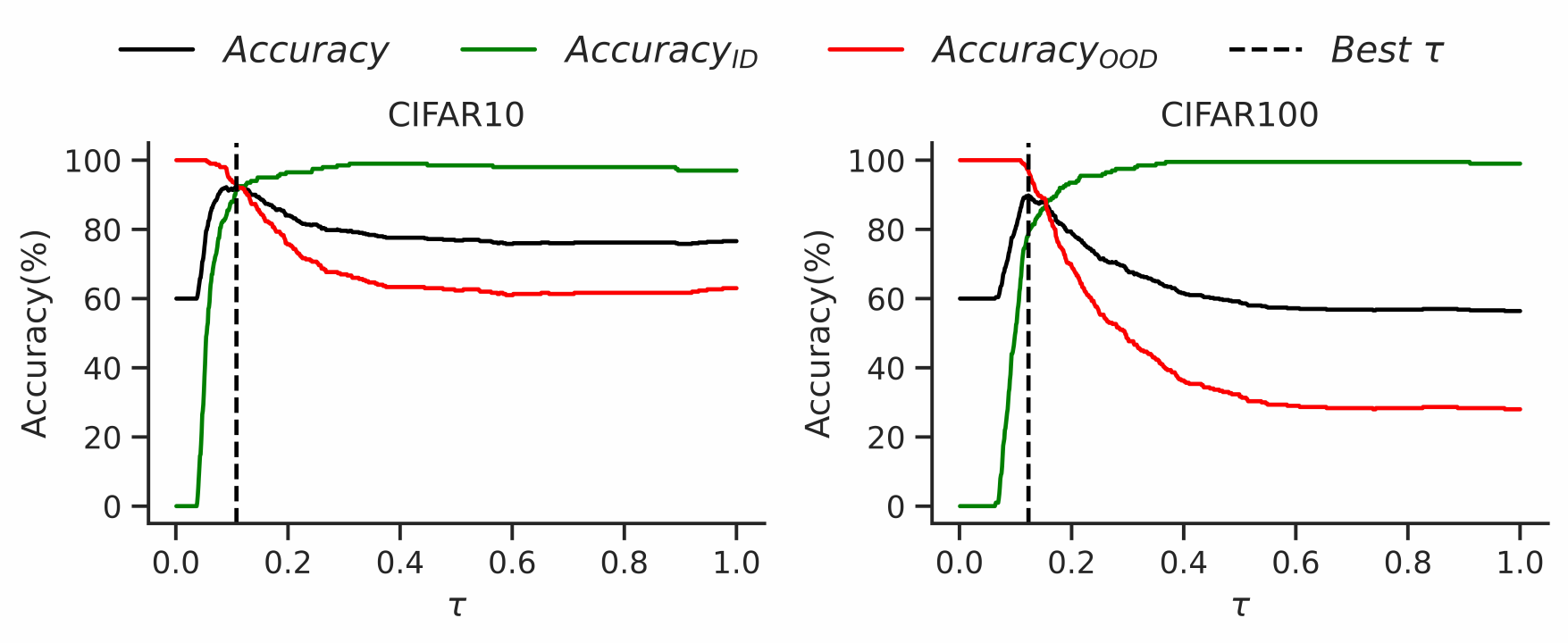}
\end{center}
\vspace{-20pt}
\caption{Changes in OOD detection accuracy of VLPure based on the temperature parameter. Both datasets are constructed with a 40\% mismatch ratio and a 60\% OOD ratio.}
\label{fig:temperature}
\end{figure}

\subsection{Self-Temperature Tuning} \label{sec:pro_temp}

The temperature parameter significantly influences model confidence in predictions. The optimal temperature varies depending on the OOD detection data domain, as shown in \cref{fig:temperature}. Labeled ID and OOD data allow for exploring the optimal temperature parameter.

\begin{equation} \label{eq:opt_temp}
\begin{aligned}
\hat{y}_i(\tau) &= \mathbb{I}^*(x_i, \tau), \\
\text{Acc}(\tau) &= \frac{1}{N_{L}}\sum_{i}^{N_{L}}{\mathbb{I}(y_i^{\text{OOD}}=\hat{y}_i(\tau))}, \\
{\tau}^{*} &= \argmax_{0 < \tau}{\text{Acc}({\tau})},
\end{aligned}    
\end{equation}

\noindent where $N_{L}$ is the number of labeled data $S = \{x_i \in S^{\text{ID}} \cup S^{\text{OOD}}\}$. $y_i^{\text{OOD}}$ indicates 1 if $x_i$ is an OOD sample and 0 otherwise.

According to \cref{eq:opt_temp}, we apply self-temperature tuning to identify the optimal temperature $\tau^{*}$ by comparing the accuracy $Acc(\tau)$ against the ground truth, $y_i^{\text{OOD}}$.

\subsection{VLPure-AL Process} \label{sec:pro_pseudo}

Algorithm \ref{algorithm:process} summarizes the entire VLPure-AL process. AL involves data collection and model training. Initially, since no labeled data exists, data is randomly selected from the unlabeled pool for model training. Given that the unlabeled data contains OOD data, the proportion of OOD data in the chosen data reflects the overall distribution. Only ID data is used for model training, excluding OOD data. 

Once initial training is complete, data is selected using the proposed VLPure-AL method. VLPure-AL assesses the purity of data from the unlabeled pool. First, it explores the optimal temperature parameter using labeled ID and OOD data, then applies CLIPN's purity assessment process and visual weighting to calculate the purity of the unlabeled data. Based on the computed purity, OOD data is excluded, and the remaining data is assessed for informativeness using a standard AL query strategy. The selected data is annotated according to predefined costs and added to the labeled data. After data collection, the classifier is trained with the newly labeled data, and the model is reinitialized and retrained upon further data collection. Once the user-defined number of rounds is complete, the final classifier trained with the last labeled data is used as the final model.

\begin{algorithm}[htbp] \label{algorithm:process}
\small
\caption{VLPure-AL Process}
\textbf{Input:} Labeled~data~$S=\{y_i, x_i \in S^{\text{ID}} \cup S^{\text{OOD}}\}_{i=1}^{N_{\text{L}}}$, Unlabeled~data~$D_\text{U}=\{x_i \in \textbf{x}^{\text{ID}} \cup \textbf{x}^{\text{OOD}} \}_{i=1}^{N_{\text{U}}}$, Query~$Q$, classifier~$f^{\text{clf}}$, Image~encoder~$f^{\text{img}}$, Text~encoder~$f^{\text{text}}$, No~text~encoder~$f^{\text{no text}}$, Temperature~$\tau$, Annotation~costs~$C$, Number~of~classes~in~labeled~data~$K$, and Round~$R$.

\textbf{Output:} $f^{\text{clf}}$

\SetKwFunction{FMain}{Purity} 
\SetKwProg{Fn}{Function}{:}{end} 

\medskip

\tcp{Initial labeled data sampling}
Randomly select samples $S=\{x_i\}_{i=1}^{C}$ from $D_\text{U}$.

Divide $S$ into $S^{\text{ID}}$ and $S^{\text{OOD}}$.

\medskip

\For{$0 \leq r \leq R$}{
    \textbf{Initialize} $f^{\text{clf}}$
    
    Train $f^{\text{clf}}$ on $S^{\text{ID}}$

    Find optimal temperature $\tau^{*}$ using Eq.\ref{eq:opt_temp}.

    $\hat{\textbf{x}}_{\text{U}}^{\text{ID}} \gets$ \FMain{$D_{\text{U}}, S^{\text{ID}}, f^{\text{img}}, f^{\text{text}}, f^{\text{no text}}, \tau^{*}$}

    $Q \gets$ StandardAL($f^{\text{clf}}, \hat{\textbf{x}}_{\text{U}}^{\text{ID}}, C$)

    Make annotation on the selected $Q$

    $S^{\text{ID}} \gets S^{\text{ID}} \cup \textbf{x}^{\text{ID}}$
    
    $S^{\text{OOD}} \gets S^{\text{OOD}} \cup \textbf{x}^{\text{OOD}}$

    $D_{\text{U}} \gets D_{\text{U}} - Q$
}
\Return $f^{\text{clf}}$

\medskip

\Fn{\FMain{$D_{\text{U}}, S_{\text{ID}}, f^{\text{img}}, f^{\text{text}}, f^{\text{no text}}, \tau$}}{ 
    \tcp{Extract the features of unlabeled data}
    Calculate $\textbf{z}^{\text{img}}$ of $D_{\text{U}}$ and $\textbf{z}^{\text{text}}$, $\textbf{z}^{\text{no text}}$ using Eq.\ref{eq:get_features}.
    \medskip
    
    \tcp{Calculate ID and OOD probability using ATD}
    Calculate $p_{ij}^{\text{ID}}$ and $p_{ij}^{\text{OOD}}$ with $\tau$, $\textbf{z}^{\text{img}}$, $\textbf{z}^{\text{text}}$, and $\textbf{z}^{\text{no text}}$ using Eq.\ref{eq:calc_prob}
    \medskip
    
    \tcp{Adjust ID and OOD probability using visual similarity weights}
    Calculate class representation $\phi_j$ of $S_{\text{ID}}$ using Eq.\ref{eq:vis_lb_vec}.
    
    Calculate visual similarity weights $s_{ij}$ with $\tau$ and $\phi_j$ using Eq.\ref{eq:vis_weights}.
    
    Adjust $p_{ij}^{\text{ID}}$ and $p_{ij}^{\text{OOD}}$ to $p_{ij}^{*, \text{ID}}$ and $p_{ij}^{*, \text{OOD}}$ using \ref{eq:update_probs}.

    \medskip
    \tcp{Select predicted ID samples}
    Select $\hat{\textbf{x}}_{\text{U}}^{\text{ID}} \gets \{\mathbb{I}^{*}(x_i)=0, x_i \in D_{\text{U}}\}$.

    \medskip
    \textbf{return} $\hat{\textbf{x}}_{\text{U}}^{\text{ID}}$
}
\end{algorithm}

\section{Experiments}
\label{sec:exp}

\subsection{Experimental Setup}

\textbf{Baselines.} To demonstrate the efficiency of VLPure-AL, we select query strategies from standard AL and open-set AL as comparison methods. For \textit{standard AL}, the following strategies are selected: (1) uncertainty-based strategy, including CONF~\citep{conf}, LL~\citep{ll}, and (2) diversity-based strategy, including CORESET~\citep{coreset}. For \textit{open-set AL}, the chosen strategies are (1) contrastive learning-based strategy MQNet~\citep{mqnet}, (2) OOD detector-based strategy, including LfOSA~\citep{lfosa} and EOAL~\citep{eoal}. To evaluate the effectiveness of query strategies, we include RANDOM, which selects data randomly without specific criteria, in the experiments. 

\begin{table}
\caption{Benchmark description.} \label{tab:benchmarks}
\vspace{5pt}
\centering
\small
\begin{adjustbox}{max width=\textwidth}
\begin{tabular}{@{}cc|cccc@{}}
\toprule
\multicolumn{2}{c|}{\textbf{Datasets}}                 & \textbf{Image Size} & \textbf{\# Classes} & \textbf{\# Trainset} & \textbf{\# Testset} \\ \midrule
\multicolumn{2}{c|}{CIFAR10}       & 32 x 32   & 10  & 50,000  & 10,000 \\
\multicolumn{2}{c|}{CIFAR100}      & 32 x 32   & 100 & 50,000  & 10,000 \\
\multicolumn{2}{c|}{Tiny-ImageNet} & 64 x 64   & 200 & 100,000 & 10,000 \\ \midrule
\multicolumn{1}{c|}{\multirow{6}{*}{DomainNet}} & real & 224 x 224           & 345                 & 120,906               & 52,041               \\
\multicolumn{1}{c|}{}  & clipart   & 224 x 224 & 345 & 33,525  & 14,604 \\
\multicolumn{1}{c|}{}  & infograph & 224 x 224 & 345 & 36,023  & 15,582 \\
\multicolumn{1}{c|}{}  & painting  & 224 x 224 & 345 & 50,416  & 21,850 \\
\multicolumn{1}{c|}{}  & sketch    & 224 x 224 & 345 & 48,212  & 20,916 \\
\multicolumn{1}{c|}{}  & quickdraw & 224 x 224 & 345 & 120,750 & 51,750 \\ \bottomrule
\end{tabular}
\end{adjustbox}
\end{table}

\begin{table}
\caption{The number of samples for open-set scenarios.} \label{tab:data_desc}
\vspace{5pt}
\centering
\small
\begin{adjustbox}{max width=\textwidth}
\begin{tabular}{@{}c|c|cccc|cccc@{}}
\toprule
\multirow{3}{*}{\textbf{\begin{tabular}[c]{@{}c@{}}Mismatch\\ Ratio\end{tabular}}} &
  \multirow{3}{*}{\textbf{\begin{tabular}[c]{@{}c@{}}OOD\\ Ratio\end{tabular}}} &
  \multicolumn{4}{c|}{\textbf{CIFAR10 \& CIFAR100}} &
  \multicolumn{4}{c}{\textbf{Tiny-imageNet}} \\ \cmidrule(l){3-10} 
 &
   &
  \multicolumn{2}{c|}{\textbf{Initial $S_L$}} &
  \multicolumn{2}{c|}{\textbf{$D_U$}} &
  \multicolumn{2}{c|}{\textbf{Initial $S_L$}} &
  \multicolumn{2}{c}{\textbf{$D_U$}} \\ \cmidrule(l){3-10} 
 &
   &
  \textbf{ID} &
  \multicolumn{1}{c|}{\textbf{OOD}} &
  \textbf{ID} &
  \textbf{OOD} &
  \textbf{ID} &
  \multicolumn{1}{c|}{\textbf{OOD}} &
  \textbf{ID} &
  \textbf{OOD} \\ \midrule
\multirow{4}{*}{20\%} & 10\% & 450 & \multicolumn{1}{c|}{50}  & 9,550  & 1,061  & 900 & \multicolumn{1}{c|}{100} & 19,100 & 2,122  \\
                      & 20\% & 400 & \multicolumn{1}{c|}{100} & 9,600  & 2,400  & 800 & \multicolumn{1}{c|}{200} & 19,200 & 4,800  \\
                      & 40\% & 300 & \multicolumn{1}{c|}{200} & 9,700  & 6,467  & 600 & \multicolumn{1}{c|}{400} & 19,400 & 12,933 \\
                      & 60\% & 200 & \multicolumn{1}{c|}{300} & 9,800  & 14,700 & 400 & \multicolumn{1}{c|}{600} & 19,600 & 29,400 \\ \midrule
\multirow{4}{*}{30\%} & 10\% & 450 & \multicolumn{1}{c|}{50}  & 14,550 & 1,617  & 900 & \multicolumn{1}{c|}{100} & 29,100 & 3,233  \\
                      & 20\% & 400 & \multicolumn{1}{c|}{100} & 14,600 & 3,650  & 800 & \multicolumn{1}{c|}{200} & 29,200 & 7,300  \\
                      & 40\% & 300 & \multicolumn{1}{c|}{200} & 14,700 & 9,800  & 600 & \multicolumn{1}{c|}{400} & 29,400 & 19,600 \\
                      & 60\% & 200 & \multicolumn{1}{c|}{300} & 14,800 & 22,200 & 400 & \multicolumn{1}{c|}{600} & 29,600 & 44,400 \\ \midrule
\multirow{4}{*}{40\%} & 10\% & 450 & \multicolumn{1}{c|}{50}  & 19,550 & 2,172  & 900 & \multicolumn{1}{c|}{100} & 39,100 & 4,344  \\
                      & 20\% & 400 & \multicolumn{1}{c|}{100} & 19,600 & 4,900  & 800 & \multicolumn{1}{c|}{200} & 39,200 & 9,800  \\
                      & 40\% & 300 & \multicolumn{1}{c|}{200} & 19,700 & 13,133 & 600 & \multicolumn{1}{c|}{400} & 39,400 & 26,267 \\
                      & 60\% & 200 & \multicolumn{1}{c|}{300} & 19,800 & 29,700 & 400 & \multicolumn{1}{c|}{600} & 39,600 & 59,400 \\ \bottomrule
\end{tabular}
\end{adjustbox}
\end{table}

\begin{table}
\caption{The number of samples for DomainNet open-set scenarios.} \label{tab:domainnet_desc}
\vspace{5pt}
\centering
\small
\begin{adjustbox}{max width=\textwidth}
\begin{tabular}{@{}c|c|cccc@{}}
\toprule
\multirow{3}{*}{\textbf{\begin{tabular}[c]{@{}c@{}}OOD\\ Ratio\end{tabular}}} & \multirow{3}{*}{\textbf{\begin{tabular}[c]{@{}c@{}}\# ID\\Class\end{tabular}}} & \multicolumn{4}{c}{\textbf{DomainNet}} \\ \cmidrule(l){3-6} 
          &    & \multicolumn{2}{c|}{\textbf{Initial $S_L$}}    & \multicolumn{2}{c}{\textbf{$D_U$}} \\ \cmidrule(l){3-6} 
          &    & \textbf{ID} & \multicolumn{1}{c|}{\textbf{OOD}} & \textbf{ID}     & \textbf{OOD}    \\ \midrule
real      & 50 & 50          & \multicolumn{1}{c|}{0}            & 15,440           & 394,342          \\
clipart   & 50 & 50          & \multicolumn{1}{c|}{0}            & 3,944            & 405,838          \\
infograph & 50 & 50          & \multicolumn{1}{c|}{0}            & 4,957            & 404,825          \\
painting  & 50 & 50          & \multicolumn{1}{c|}{0}            & 6,806            & 400,816          \\
sketch    & 50 & 50          & \multicolumn{1}{c|}{0}            & 6,218            & 403,564          \\
quickdraw & 50 & 50          & \multicolumn{1}{c|}{0}            & 17,450           & 392,332          \\ \bottomrule
\end{tabular}
\end{adjustbox}
\end{table}

\textbf{Open-set Data.} We perform experiments on CIFAR10, CIFAR100~\citep{cifar}, Tiny-ImageNet~\citep{tiny}, and DomainNet~\citep{domainnet}, with detailed descriptions of the benchmarks provided in \cref{tab:benchmarks}. To validate the effectiveness of open-set AL, we reconstruct the datasets by employing mismatch and OOD ratios for open-set data, with the detailed configurations provided in \cref{tab:data_desc}. The mismatch ratio represents the proportion of ID classes in the dataset~\citep{lfosa, eoal}, while the OOD ratio denotes the proportion of OOD data in the unlabeled pool~\citep{ccal, mqnet}. Previous experiments by LfOSA and EOAL focused on open-set data constructed with the mismatch ratio but did not consider the OOD ratio. We incorporate both ratios to create more realistic scenarios when constructing open-set data. For instance, in CIFAR10 with a mismatch ratio of 20\%, two classes are ID, and the remaining eight are OOD. The mismatch ratios are defined as 20\%, 30\%, and 40\%. The OOD ratios are defined as 10\%, 20\%, 40\%, and 60\%. The composition of the open-set data used in the experiments is detailed in \cref{tab:data_desc}. 

To simulate scenarios where different domains are included in the unlabeled data, we constructed an open-set configuration using the DomainNet dataset, with the detailed configurations provided in \cref{tab:domainnet_desc}. Among the six domains in DomainNet, one domain is designated as in-domain, while the remaining five are treated as out-of-domain. Additionally, out of the 345 classes in each domain, 50 classes are selected as in-distribution, and the rest are considered out-of-distribution.

We evaluate the robustness and effectiveness of the query strategies under various scenarios. The annotation cost $C$ for AL is 500 for CIFAR10 and CIFAR100, 1000 for Tiny-ImageNet, and 50 for DomainNet, where DomainNet includes one image per class. The initial data for training the classifier is randomly selected from the unlabeled pool. In subsequent iterations, data is selected based on the query strategy with a cost of $C$.

\textbf{Implementation Details.} The image classifier is ResNet18~\citep{resnet} in the experiments. The number of rounds $R$ for AL experiments is set to 10. Each query strategy experiment is repeated five times, with average results reported. 

The experiments use ResNet18~\citep{resnet} as the image classifier for the CIFAR10, CIFAR100, and Tiny-ImageNet datasets. The model is trained for 20,000 steps with a batch size of 64 for all datasets. The learning rate was set to 0.1, weight decay to 0.0005, and momentum to 0.9, with the model parameters updated using Stochastic Gradient Descent (SGD). 

For the experiments on DomainNet, we use the ViT-B/16~\citep{vit} pre-trained on ImageNet-21k. The pre-trained model weights are kept frozen, and only a linear classifier is trained. The model is trained for 100 epochs with a learning rate of 0.01, weight decay of 0.0005, and momentum of 0.9, with model parameters updated using Stochastic Gradient Descent (SGD).

The image encoder and text encoder for CLIPN were pre-trained on LAION-2B~\citep{openclip}, and the \textit{"no"} text encoder was pre-trained on CC3M~\citep{cc3m}. The image encoder uses a ViT-B/16 with a patch size of $16 \times 16$. For data informativeness, VLPure-AL uses CONF~\citep{conf}, one of the query strategies proposed in standard AL. Hyper-parameters for LL, MQNet, LfOSA, and EOAL are set as per the original studies~\citep{ll, mqnet, lfosa, eoal}. Experiments are conducted using Python 3.8.10 and PyTorch 1.14 on an Intel Xeon CPU E5-2698 v4 with a Tesla V100 GPU.

\begin{figure*}[t!] 
\begin{center}
\includegraphics[width=1\linewidth]{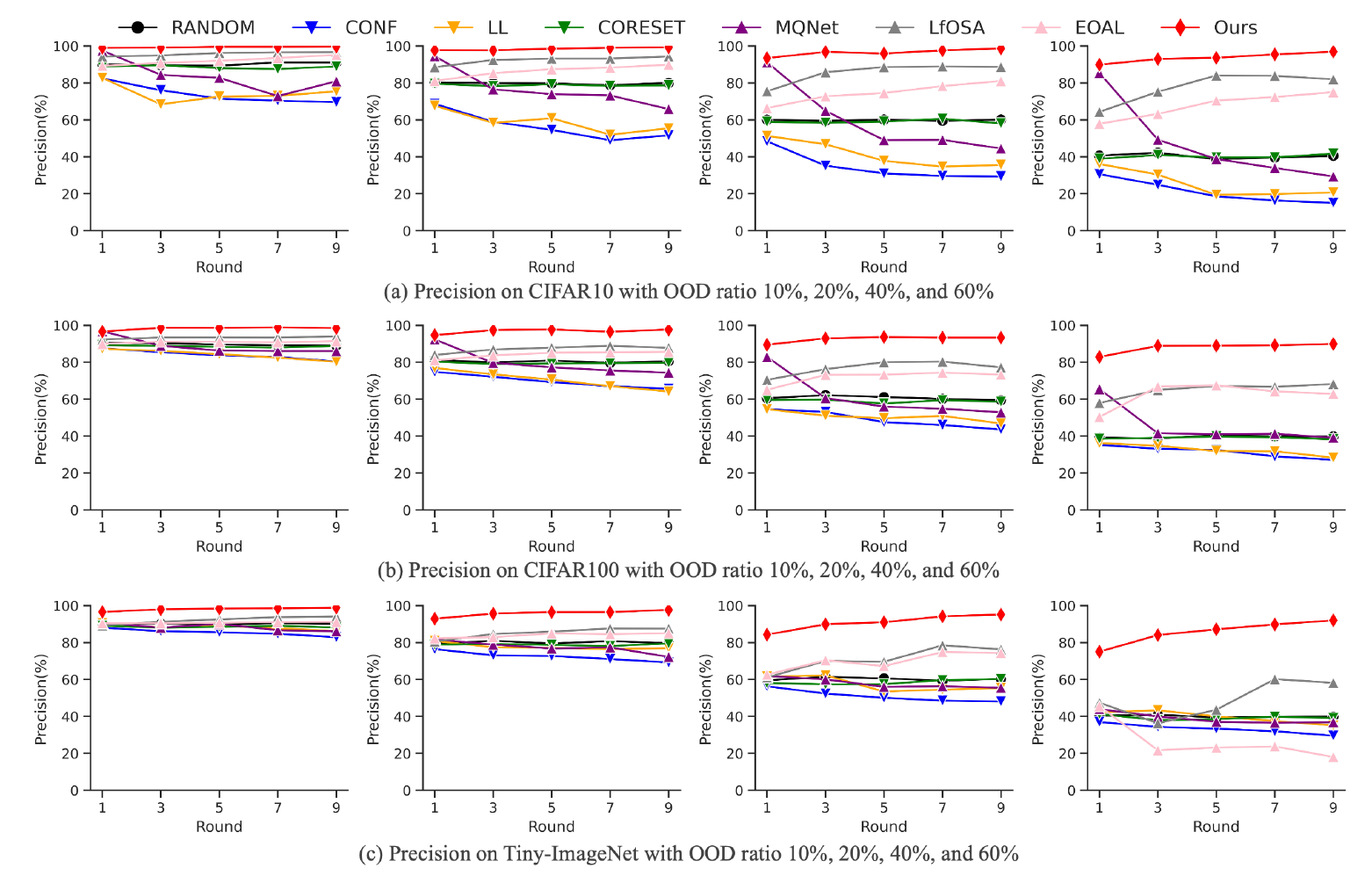}
\end{center}
\vspace{-20pt}
\caption{Comparison of precision changes in open-set AL based on OOD ratio. The mismatch ratio is set to 40\%. Precision refers to the proportion of ID data in selected queries.}
\label{fig:precision_ood}
\end{figure*}

\textbf{Evaluation Metrics.} Precision, the proportion of selected ID queries, is used to evaluate annotation cost loss in open-set AL~\citep{lfosa}. Higher precision indicates a higher proportion of ID data in the selected queries.

\begin{align}
Precision(\%) = \frac{\sum_{i=1}^{C}{\mathbb{I}(x_i = \text{ID})}}{C} \times 100.
\end{align}

AL results are assessed using Area Under Budget Curve (AUBC)~\citep{aubc}. $\text{AUBC}_{acc}$ measures the efficiency of AL by calculating the area under the curve of accuracy improvement over rounds. A larger area indicates greater accuracy improvement after data collection.

\subsection{Open-set Robustness}

\textbf{Comparison of Annotation Cost Loss.} We examine annotation cost loss through precision changes during data collection in \cref{fig:precision_ood}. Each round's precision indicates the proportion of ID data among the selected data at the round. RANDOM, shown in black, selects data randomly, so the proportion of OOD data in the selected queries matches the OOD ratio in the unlabeled pool. A higher precision than RANDOM indicates a higher selection of ID data than the OOD ratio.

\begin{figure*}[t] 
\begin{center}
\includegraphics[width=1\linewidth]{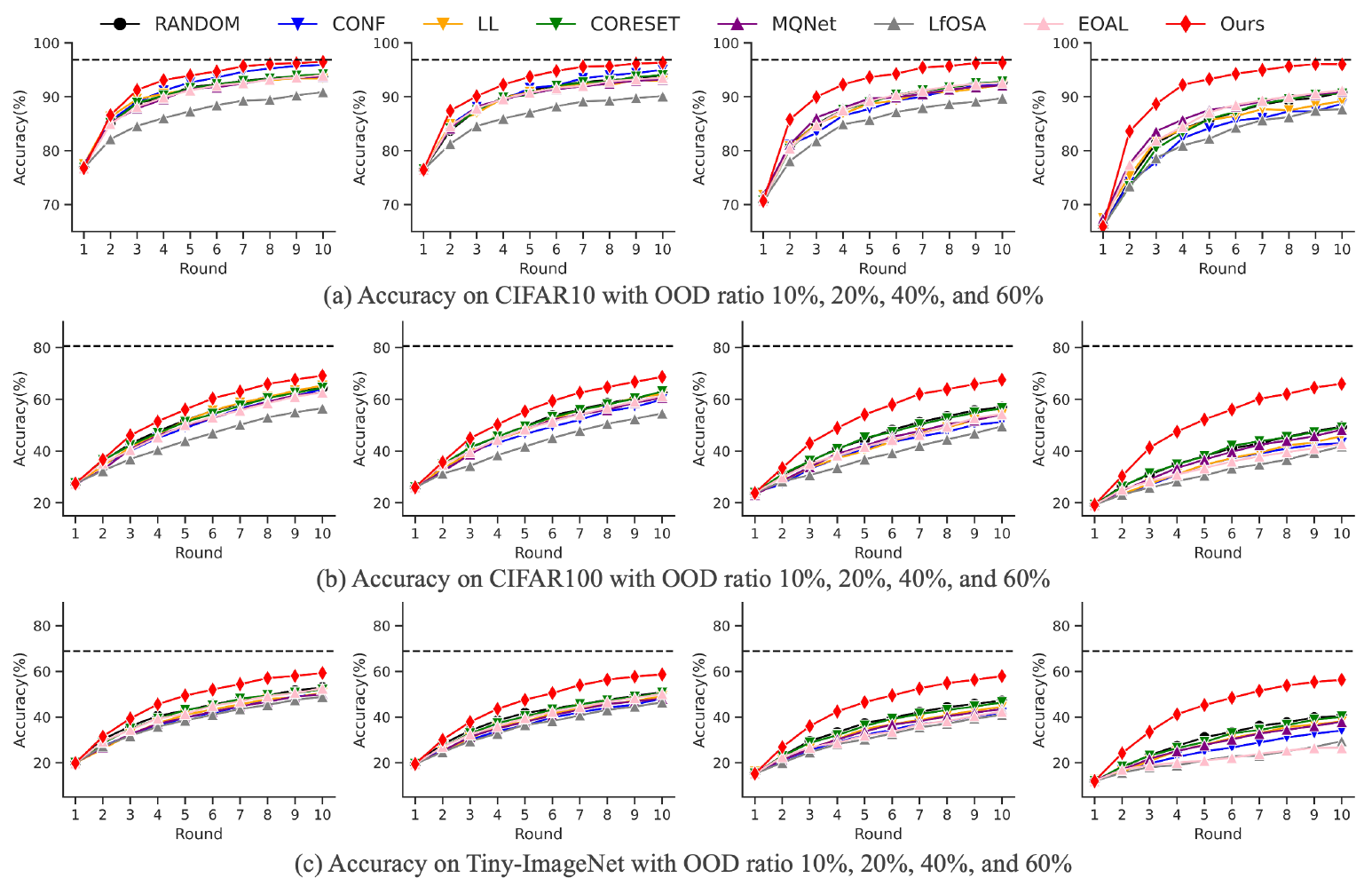}
\end{center}
\vspace{-20pt}
\caption{Comparison of accuracy changes in open-set AL based on OOD ratio. The mismatch ratio is set to 40\%. The black dashed line represents the performance of a classifier trained on the entire dataset.}
\label{fig:acc_ood}
\end{figure*}

Standard AL strategies, except CORESET, show lower precision than RANDOM in all cases, particularly as the OOD ratio increases. This indicates significant cost loss when OOD data is in the unlabeled pool. In contrast, most open-set AL strategies show higher precision than RANDOM. However, MQNet, LfOSA, and EOAL do not show consistent precision across rounds. MQNet initially focuses on purity, selecting more ID data, but later prioritizes informativeness, causing precision to drop. LfOSA and EOAL initially had low precision due to the limited OOD data for training the detection model, but they improved as more OOD data was collected. Unlike existing strategies, VLPure-AL maintains high precision throughout data collection, regardless of the OOD ratio, indicating minimal annotation cost loss.

\begin{figure*}[t!]
\begin{center}
\includegraphics[width=1\linewidth]{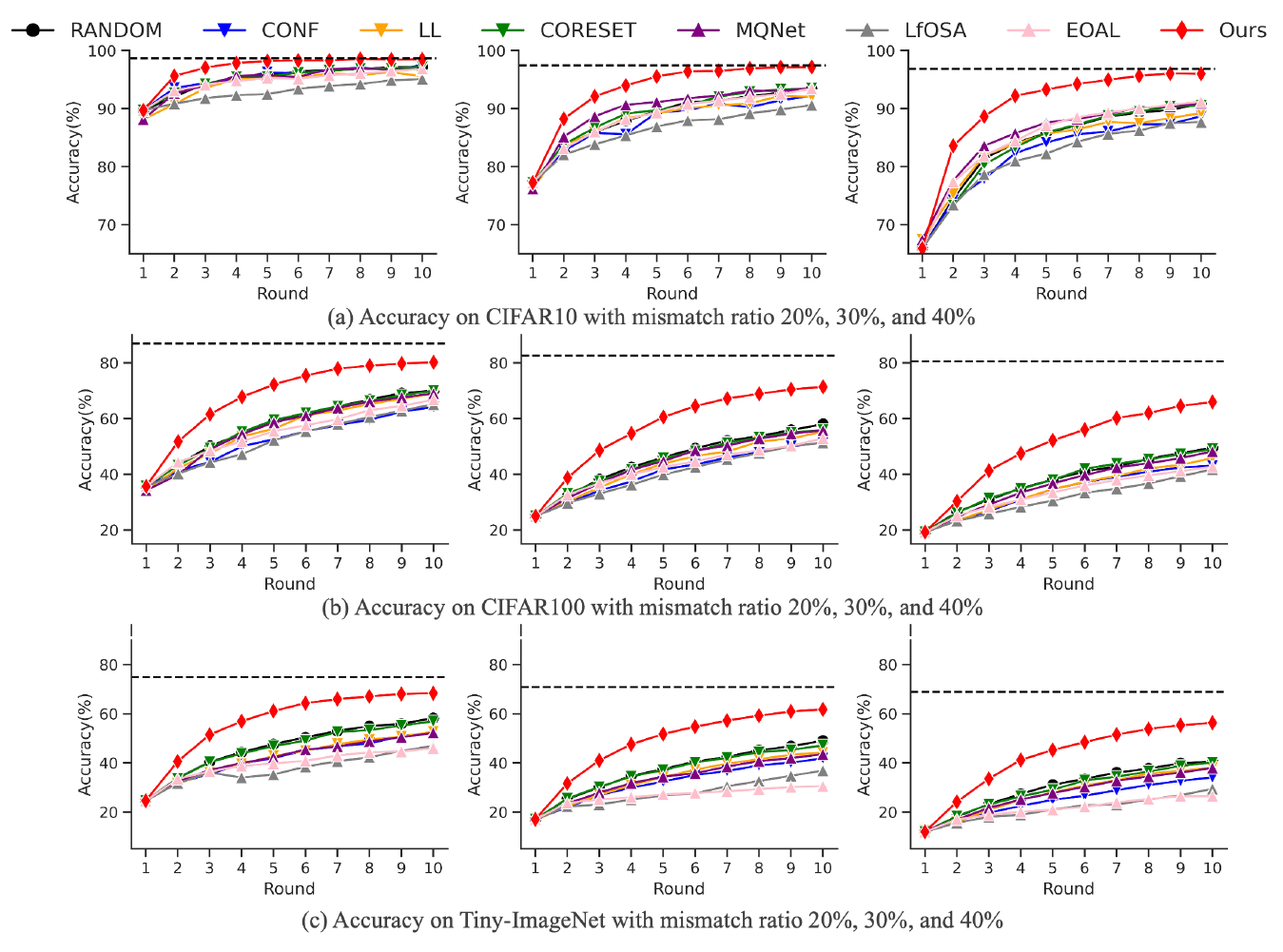}
\end{center}
\vspace{-20pt}
\caption{Comparison of accuracy changes in open-set AL based on mismatch ratio. The OOD ratio is set to 60\%. The black dashed line represents the performance of a classifier trained on the entire dataset.}
\label{fig:acc_id}
\end{figure*}

\textbf{Robustness to OOD Ratio.} \cref{fig:acc_ood} shows the results for (a) CIFAR10, (b) CIFAR100, and (c) Tiny-ImageNet with varying OOD ratios. The closer the model performance is to the black dashed line, the more informative the labeled ID data. VLPure-AL demonstrates robust performance, with minimal accuracy drop at the final round regardless of the OOD ratio. In CIFAR10, the accuracy of the classifier trained with ten rounds approximates that of a classifier trained on the entire dataset. This indicates that VLPure-AL minimizes cost loss and selects highly informative data.

Standard and open-set AL methods show varying results depending on the OOD ratio. For OOD ratios of 10\% and 20\%, standard AL methods perform better than open-set methods on CIFAR10, suggesting that standard AL methods select more informative data when the proportion of OOD data is low. However, as the OOD ratio increases, open-set AL methods outperform standard AL methods, as they can collect more ID data, enhancing classifier accuracy. LfOSA performs poorly across all cases, indicating its inability to select highly informative data despite high purity.

\textbf{Robustness to Mismatch Ratio.} \cref{fig:acc_id} shows the results for (a) CIFAR10, (b) CIFAR100, and (c) Tiny-ImageNet with varying mismatch ratios. VLPure-AL shows significant accuracy improvement in the initial data collection iterations compared to other query strategies. CIFAR10's performance remains close to the classifier trained on the entire dataset regardless of the mismatch ratio. In CIFAR100 and Tiny-ImageNet, despite a larger number of classes, VLPure-AL maintains high accuracy compared to other strategies. 

\begin{table}
\caption{$\text{AUBC}_{acc}$(\%) performance of query strategies based on the mismatch and OOD ratio.} \label{tab:acc_aubc}
\vspace{5pt}
\centering
\small
\begin{adjustbox}{max width=\textwidth}
\begin{tabular}{@{}c|l|cccc|cccc|cccc@{}}
\toprule
\multirow{2}{*}{\textbf{\begin{tabular}[c]{@{}c@{}}Mismatch\\ Ratio\end{tabular}}} &
  \textbf{Datasets} &
  \multicolumn{4}{c|}{\textbf{CIFAR10}} &
  \multicolumn{4}{c|}{\textbf{CIFAR100}} &
  \multicolumn{4}{c}{\textbf{Tiny-ImageNet}} \\ \cmidrule(l){2-14} 
 &
  \textbf{OOD ratio} &
  \textbf{10\%} &
  \textbf{20\%} &
  \textbf{40\%} &
  \textbf{60\%} &
  \textbf{10\%} &
  \textbf{20\%} &
  \textbf{40\%} &
  \textbf{60\%} &
  \textbf{10\%} &
  \textbf{20\%} &
  \textbf{40\%} &
  \textbf{60\%} \\ \midrule
\multirow{8}{*}{20\%} &
  RANDOM &
  87.40 &
  87.23 &
  86.78 &
  85.79 &
  63.01 &
  61.89 &
  57.51 &
  {\ul 52.13} &
  {\ul 52.62} &
  {\ul 51.03} &
  {\ul 47.75} &
  {\ul 42.17} \\ \cmidrule(l){2-14} 
 &
  CONF &
  {\ul 88.09} &
  {\ul 87.80} &
  {\ul 86.91} &
  {\ul 85.92} &
  63.22 &
  61.06 &
  54.83 &
  47.31 &
  51.24 &
  48.98 &
  44.10 &
  37.79 \\
 &
  LL &
  86.74 &
  86.66 &
  85.92 &
  85.03 &
  {\ul 64.72} &
  {\ul 63.01} &
  {\ul 57.80} &
  50.74 &
  51.63 &
  49.23 &
  44.36 &
  38.12 \\
 &
  CORESET &
  87.29 &
  87.23 &
  86.49 &
  85.79 &
  62.93 &
  61.57 &
  57.73 &
  52.10 &
  52.22 &
  49.88 &
  46.82 &
  41.53 \\ \cmidrule(l){2-14} 
 &
  MQNet &
  86.61 &
  86.56 &
  86.20 &
  85.63 &
  62.68 &
  61.19 &
  56.66 &
  51.19 &
  50.40 &
  48.58 &
  44.06 &
  38.15 \\
 &
  LfOSA &
  85.50 &
  85.01 &
  84.57 &
  83.62 &
  58.73 &
  57.46 &
  53.27 &
  47.10 &
  50.02 &
  48.10 &
  44.14 &
  33.90 \\
 &
  EOAL &
  86.97 &
  86.59 &
  86.29 &
  85.39 &
  62.70 &
  61.44 &
  56.59 &
  49.62 &
  51.99 &
  49.76 &
  45.90 &
  35.64 \\ \cmidrule(l){2-14} 
 &
  Ours &
  \textbf{88.10} &
  \textbf{88.03} &
  \textbf{87.84} &
  \textbf{87.64} &
  \textbf{65.83} &
  \textbf{65.36} &
  \textbf{64.01} &
  \textbf{62.33} &
  \textbf{55.90} &
  \textbf{55.17} &
  \textbf{53.81} &
  \textbf{52.17} \\ \midrule
\multirow{8}{*}{30\%} &
  RANDOM &
  83.80 &
  83.28 &
  {\ul 82.27} &
  79.87 &
  53.22 &
  51.35 &
  {\ul 47.00} &
  {\ul 41.14} &
  {\ul 43.84} &
  {\ul 42.56} &
  {\ul 38.69} &
  {\ul 33.51} \\ \cmidrule(l){2-14} 
 &
  CONF &
  {\ul 84.87} &
  {\ul 84.00} &
  82.01 &
  79.03 &
  52.19 &
  49.59 &
  43.55 &
  36.83 &
  41.90 &
  39.79 &
  34.93 &
  29.11 \\
 &
  LL &
  83.78 &
  83.09 &
  81.49 &
  79.50 &
  {\ul 54.07} &
  {\ul 51.42} &
  45.38 &
  38.87 &
  42.12 &
  40.51 &
  35.80 &
  30.64 \\
 &
  CORESET &
  83.65 &
  83.28 &
  82.13 &
  80.35 &
  53.12 &
  50.88 &
  46.79 &
  40.59 &
  43.23 &
  41.88 &
  38.15 &
  33.10 \\ \cmidrule(l){2-14} 
 &
  MQNet &
  83.46 &
  83.22 &
  82.19 &
  {\ul 81.05} &
  52.18 &
  50.29 &
  46.02 &
  40.10 &
  41.56 &
  40.09 &
  35.58 &
  30.42 \\
 &
  LfOSA &
  80.90 &
  80.76 &
  79.59 &
  77.72 &
  48.10 &
  45.93 &
  41.48 &
  36.24 &
  40.41 &
  38.67 &
  33.94 &
  24.93 \\
 &
  EOAL &
  83.24 &
  82.94 &
  81.86 &
  79.86 &
  52.40 &
  50.57 &
  45.76 &
  38.13 &
  43.10 &
  41.26 &
  35.57 &
  24.12 \\ \cmidrule(l){2-14} 
 &
  Ours &
  \textbf{85.50} &
  \textbf{85.33} &
  \textbf{85.04} &
  \textbf{84.40} &
  \textbf{56.00} &
  \textbf{55.34} &
  \textbf{53.75} &
  \textbf{52.17} &
  \textbf{47.99} &
  \textbf{47.50} &
  \textbf{46.12} &
  \textbf{44.32} \\ \midrule
\multirow{8}{*}{40\%} &
  RANDOM &
  81.34 &
  80.86 &
  78.96 &
  75.89 &
  {\ul 46.20} &
  {\ul 44.36} &
  {\ul 40.08} &
  34.16 &
  {\ul 38.02} &
  {\ul 36.35} &
  {\ul 32.60} &
  {\ul 27.28} \\ \cmidrule(l){2-14} 
 &
  CONF &
  {\ul 82.38} &
  {\ul 81.35} &
  78.24 &
  74.20 &
  44.38 &
  41.88 &
  36.25 &
  30.65 &
  35.25 &
  33.08 &
  28.59 &
  22.59 \\
 &
  LL &
  81.47 &
  80.66 &
  78.56 &
  75.50 &
  45.84 &
  43.56 &
  36.94 &
  31.09 &
  35.72 &
  34.14 &
  30.16 &
  25.09 \\
 &
  CORESET &
  81.35 &
  80.81 &
  78.90 &
  75.70 &
  45.72 &
  44.27 &
  39.81 &
  {\ul 34.27} &
  37.45 &
  35.70 &
  31.85 &
  26.52 \\ \cmidrule(l){2-14} 
 &
  MQNet &
  81.06 &
  80.60 &
  {\ul 79.03} &
  {\ul 77.15} &
  44.33 &
  42.75 &
  37.72 &
  32.94 &
  35.34 &
  33.94 &
  29.80 &
  24.99 \\
 &
  LfOSA &
  78.13 &
  77.85 &
  76.35 &
  73.56 &
  40.05 &
  38.18 &
  33.86 &
  28.24 &
  34.47 &
  32.28 &
  27.54 &
  19.17 \\
 &
  EOAL &
  81.08 &
  80.74 &
  78.93 &
  76.78 &
  44.43 &
  43.22 &
  37.68 &
  30.33 &
  37.19 &
  35.13 &
  28.76 &
  19.37 \\ \cmidrule(l){2-14} 
 &
  Ours &
  \textbf{83.41} &
  \textbf{83.20} &
  \textbf{82.67} &
  \textbf{81.95} &
  \textbf{49.52} &
  \textbf{48.67} &
  \textbf{47.47} &
  \textbf{45.65} &
  \textbf{42.67} &
  \textbf{41.64} &
  \textbf{40.14} &
  \textbf{38.72} \\ \bottomrule
\end{tabular}
\end{adjustbox}
\end{table}

\textbf{Open-set Robustness.} \cref{tab:acc_aubc} summarizes the $\text{AUBC}_{acc}$ results for all open-set scenarios. MQNet shows high $\text{AUBC}_{acc}$ in CIFAR10 with high OOD and mismatch ratios but does not outperform standard AL in all cases. In CIFAR100 and Tiny-ImageNet, RANDOM often outperforms other strategies. However, VLPure-AL consistently shows higher $\text{AUBC}_{acc}$ than existing methods, indicating significant accuracy improvement from the initial round.

\begin{table}
\caption{$\text{AUBC}_{acc}$(\%) and Last ACC(\%) performance of query strategies on DomainNet. "w/ domain" indicates that domain information is provided in the prompt format "A photo of \{domain\} \{class\}."}\label{tab:domainnet_aubc}
\vspace{5pt}
\centering
\small
\begin{adjustbox}{max width=\textwidth}
\begin{tabular}{@{}c|cccccc|cccccc@{}}
\toprule
\multirow{2}{*}{Methods} & \multicolumn{6}{c|}{$\text{AUBC}_{acc}$}                                             & \multicolumn{6}{c}{Last ACC}                                     \\ \cmidrule(l){2-13} 
                         & real           & clipart & infograph & painting & sketch & quickdraw      & real  & clipart & infograph & painting & sketch & quickdraw      \\ \midrule
RANDOM                   & 60.76          & 28.31   & 20.39     & 38.20    & 30.75  & {\ul 29.34}          & 70.07 & 32.03   & 23.78     & 43.56    & 35.55  & {\ul 35.18}          \\ \midrule
CONF                     & 59.29          & 28.74   & 20.13     & 36.50    & 29.68  & 27.14          & 68.18 & 32.34   & 23.05     & 41.12    & 33.52  & 29.65          \\
LL                       & 61.41          & 31.75   & 23.15     & 43.70    & 31.43  & 27.22          & 69.57 & 37.23   & 28.80     & {\ul 52.36}    & 36.03  & 30.62          \\ \midrule
LfOSA                    & 61.31          & 30.71   & 22.28     & 41.63    & 30.85  & \textbf{29.54} & 65.84 & 38.21   & {\ul 29.16}     & 49.46    & 38.51  & \textbf{36.43} \\
MQNet                    & 58.80          & {\ul 32.12}   & 22.97     & {\ul 44.19}    & {\ul 32.81}  & 27.74          & 61.26 & {\ul 40.92}   & 27.99     & 52.24    & {\ul 39.31}  & 33.03          \\ \midrule
Ours                  & \textbf{70.59} & 32.03   & {\ul 24.32}     & 43.46    & 32.38  & 26.75          & {\ul 79.96} & 37.62   & 28.03     & 50.76    & 37.34  & 29.71          \\
Ours w/ domain &
  {\ul 70.16} &
  \textbf{35.53} &
  \textbf{25.73} &
  \textbf{47.88} &
  \textbf{37.51} &
  27.13 &
  \textbf{80.26} &
  \textbf{42.08} &
  \textbf{30.27} &
  \textbf{57.51} &
  \textbf{44.96} &
  30.16 \\ \bottomrule
\end{tabular}
\end{adjustbox}
\end{table}

\textbf{Domain Robustness.} \cref{tab:domainnet_aubc} shows the experimental results for Open-set AL when different classes and domains are mixed as OOD data. In the case of Open-set DomainNet, the unlabeled pool contains more complex OOD samples compared to CIFAR10, CIFAR100, and Tiny-ImageNet. As a result, Open-set AL methods generally outperform not only RANDOM but also Standard AL. For VLPure-AL, a significant performance improvement is observed when text information providing domain context (e.g., "A photo of \{domain\} \{class\}") is used compared to cases where such information is not provided. Moreover, except for the quickdraw domain, VLPure-AL consistently achieves higher $\text{AUBC}_{acc}$ and Last ACC in almost all cases, highlighting the advantage of incorporating domain information depending on the scenario.

\begin{figure}
\begin{center}
\includegraphics[width=1\linewidth]{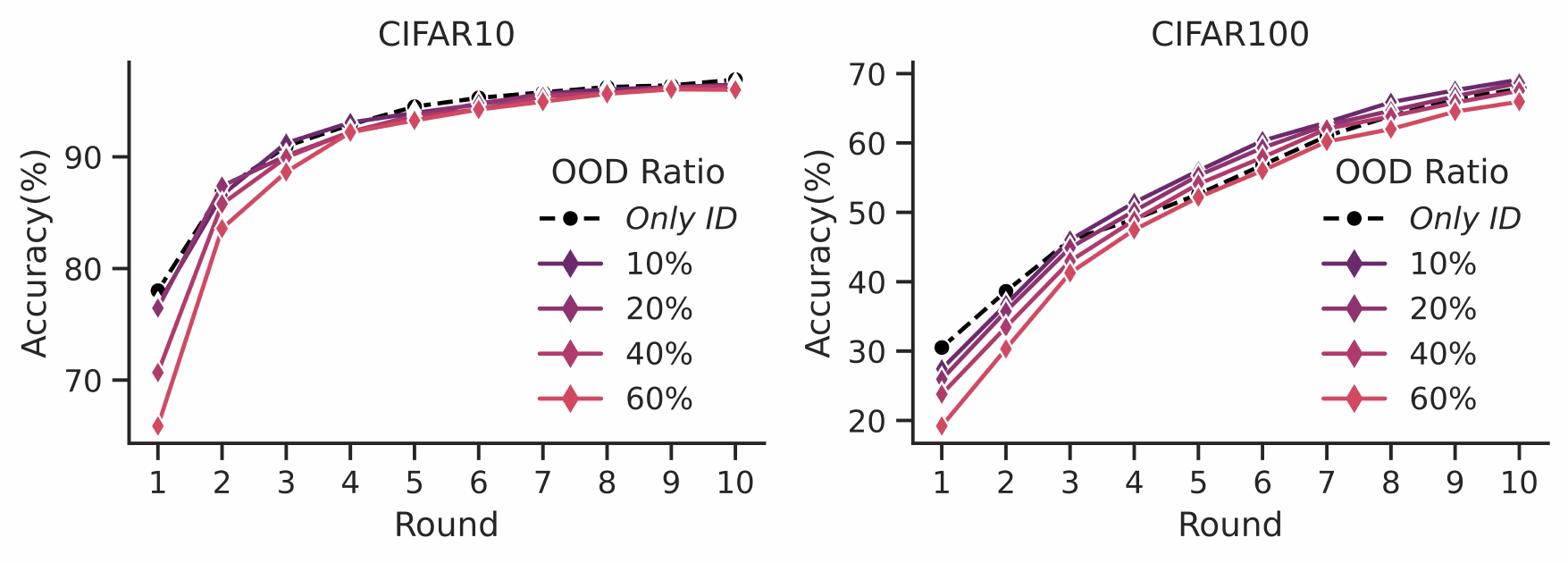}
\end{center}
\vspace{-20pt}
\caption{The robustness of VLPure-AL. The mismatch ratio is 40\% for both datasets.}
\label{fig:ours_robust}
\end{figure}

\textbf{Comparison with Standard AL Scenarios.} \cref{fig:ours_robust} demonstrates the robustness of VLPure-AL. The black dashed line represents the accuracy change during data collection assuming no OOD data, applying CONF to only ID data. Without OOD data, there is no cost loss. VLPure-AL, despite the presence of OOD data, shows accuracy convergence to the black dashed line in CIFAR10 and even surpasses it in CIFAR100. This indicates that VLPure-AL selects more informative data than standard AL assuming only ID data.

\begin{table}
\caption{Comparison of zero-shot, full-shot, and VLPure-AL performance. Zero-shot represents the performance of CLIP, while full-shot refers to the performance achieved by training only on ID samples. Both full-shot and the classifier used for active learning utilize CLIP's visual encoder.} \label{tab:zeroshot}
\vspace{5pt}
\centering
\small
\begin{adjustbox}{max width=\textwidth}
\begin{tabular}{@{}c|c|cccccc@{}}
\toprule
\multirow{2}{*}{\textbf{Datasets}} &
  \multirow{2}{*}{\textbf{\begin{tabular}[c]{@{}c@{}}Mismatch\\ Ratio\end{tabular}}} &
  \multicolumn{6}{c}{\textbf{Accuracy   (\%)}} \\ \cmidrule(l){3-8} 
 &
   &
  \multicolumn{1}{c|}{\textbf{zero-shot}} &
  \textbf{round 0} &
  \textbf{round 1} &
  \textbf{round 2} &
  \multicolumn{1}{c|}{\textbf{round 3}} &
  \textbf{full-shot} \\ \midrule
\multirow{3}{*}{CIFAR10} &
  20\% &
  \multicolumn{1}{c|}{99.76} &
  99.76 &
  99.75 &
  \textbf{99.80} &
  \multicolumn{1}{c|}{99.83} &
  99.77 \\
 &
  30\% &
  \multicolumn{1}{c|}{99.06} &
  \textbf{99.15} &
  99.21 &
  99.43 &
  \multicolumn{1}{c|}{99.37} &
  99.32 \\
 &
  40\% &
  \multicolumn{1}{c|}{98.54} &
  \textbf{98.75} &
  98.88 &
  99.14 &
  \multicolumn{1}{c|}{99.16} &
  99.04 \\ \midrule
\multirow{3}{*}{CIFAR100} &
  20\% &
  \multicolumn{1}{c|}{90.90} &
  \textbf{91.19} &
  93.04 &
  93.80 &
  \multicolumn{1}{c|}{93.97} &
  94.29 \\
 &
  30\% &
  \multicolumn{1}{c|}{87.67} &
  85.96 &
  \textbf{87.59} &
  89.53 &
  \multicolumn{1}{c|}{90.35} &
  91.41 \\
 &
  40\% &
  \multicolumn{1}{c|}{84.31} &
  80.62 &
  83.14 &
  \textbf{85.33} &
  \multicolumn{1}{c|}{85.88} &
  89.31 \\ \midrule
\multirow{3}{*}{Tiny-ImageNet} &
  20\% &
  \multicolumn{1}{c|}{82.79} &
  81.62 &
  \textbf{84.85} &
  86.65 &
  \multicolumn{1}{c|}{86.99} &
  88.70 \\
 &
  30\% &
  \multicolumn{1}{c|}{79.89} &
  76.30 &
  \textbf{80.17} &
  82.05 &
  \multicolumn{1}{c|}{83.14} &
  85.90 \\
 &
  40\% &
  \multicolumn{1}{c|}{77.98} &
  71.92 &
  74.99 &
  77.44 &
  \multicolumn{1}{c|}{\textbf{78.50}} &
  83.45 \\ \bottomrule
\end{tabular}
\end{adjustbox}
\end{table}

\textbf{Comparison AL with zero-shot and full-shot.} \cref{tab:zeroshot} shows the difference between zero-shot and full-shot performance when applying active learning to open-set data. Full-shot refers to the performance achieved by training solely on ID samples. The classifier used for active learning employed the visual encoder from CLIP, the same one used for the zero-shot results, to allow for a direct comparison. While zero-shot performance using CLIP is generally strong compared to full-shot, the results demonstrate that collecting data through active learning can surpass zero-shot performance, especially when the mismatch ratio is small, requiring only a small number of samples. Furthermore, VLPure-AL rapidly reaches full-shot performance within just four rounds, even in open-set data scenarios.

\subsection{Ablation Studies} \label{sec:ablation}

\begin{figure}[t]
\begin{center}
\includegraphics[width=1\linewidth]{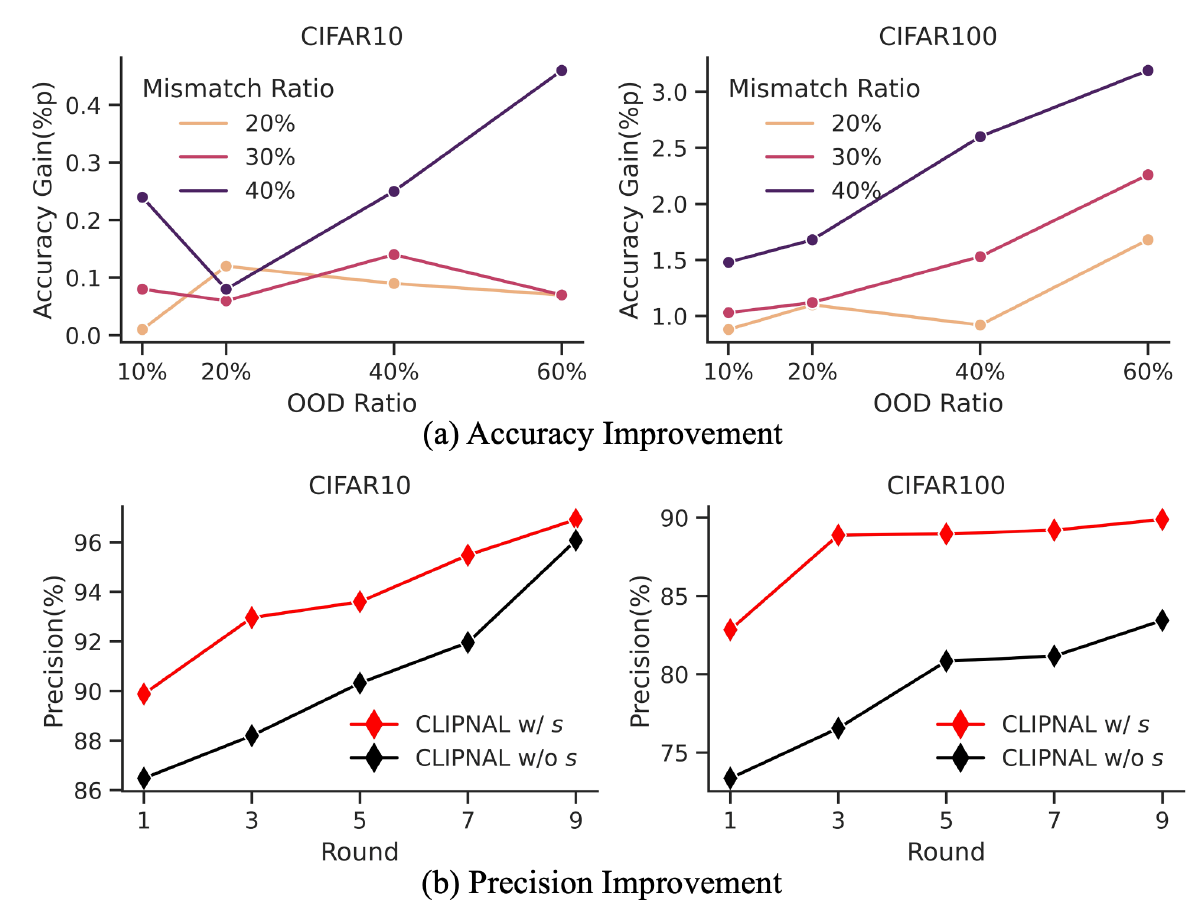}
\end{center}
\vspace{-20pt}
\caption{Changes in accuracy and precision with and without visual similarity weighting. Open-set data for (b) precision improvement is constructed a 40\% mismatch ratio and a 60\% OOD ratio.}
\label{fig:acc_precision_gain}
\end{figure}

\textbf{Effect of Visual Similarity Weights.} To improve the accuracy of OOD detection, VLPure-AL uses visual similarity weights from labeled ID data. \cref{fig:acc_precision_gain} shows (a) accuracy and (b) precision improvements with and without visual similarity weights. (a) Accuracy improvement is significant with increasing mismatch ratios and OOD ratios, except for a 20\% OOD ratio in CIFAR10. (b) Precision results indicate that visual similarity weights significantly increase the amount of ID data selected, especially in CIFAR100, demonstrating the effectiveness of visual similarity weights in preventing cost loss.

\begin{table}
\caption{$\text{AUBC}_{acc}$(\%) and performance differences (\%p) in an ablation study analyzing the effects of Self-Temperature Tuning and Visual Similarity Weights.}\label{tab:ablation_components}
\vspace{5pt}
\centering
\scriptsize
\begin{adjustbox}{max width=\textwidth}
\begin{tabular}{@{}cc|cccc|cccc|cccc@{}}
\toprule
 &
   &
  \multicolumn{4}{c|}{\textbf{CIFAR10}} &
  \multicolumn{4}{c|}{\textbf{CIFAR100}} &
  \multicolumn{4}{c}{\textbf{Tiny-ImageNet}} \\ \cmidrule(l){3-14} 
\multirow{-2}{*}{\textbf{\begin{tabular}[c]{@{}c@{}}Self-Temperature\\ Tuning\end{tabular}}} &
  \multirow{-2}{*}{\textbf{\begin{tabular}[c]{@{}c@{}}Visual Similarity\\ weights\end{tabular}}} &
  \multicolumn{2}{c}{\textbf{10\%}} &
  \multicolumn{2}{c|}{\textbf{60\%}} &
  \multicolumn{2}{c}{\textbf{10\%}} &
  \multicolumn{2}{c|}{\textbf{60\%}} &
  \multicolumn{2}{c}{\textbf{10\%}} &
  \multicolumn{2}{c}{\textbf{60\%}} \\ \midrule
\multicolumn{2}{c|}{Best $\text{AUBC}_{acc}$ among baselines} &
  82.38 &
  - &
  77.15 &
  - &
  46.20 &
  - &
  34.27 &
  - &
  38.02 &
  - &
  27.28 &
  - \\ \midrule
\multicolumn{2}{c|}{Ours} &
  83.41 &
  - &
  81.95 &
  - &
  49.52 &
  - &
  45.65 &
  - &
  42.67 &
  - &
  38.72 &
  - \\ \midrule
\multicolumn{1}{l}{} &
  \ding{51} &
  83.37 &
  {\color[HTML]{FF0000} (-0.04)} &
  81.65 &
  {\color[HTML]{FF0000} (-0.30)} &
  48.29 &
  {\color[HTML]{FF0000} (-1.23)} &
  46.11 &
  {\color[HTML]{3C7D22} (+0.46)} &
  41.12 &
  {\color[HTML]{FF0000} (-1.55)} &
  38.45 &
  {\color[HTML]{FF0000} (-0.27)} \\
\ding{51} &
  \multicolumn{1}{l|}{} &
  83.17 &
  {\color[HTML]{FF0000} (-0.24)} &
  81.49 &
  {\color[HTML]{FF0000} (-0.46)} &
  48.04 &
  {\color[HTML]{FF0000} (-1.48)} &
  42.46 &
  {\color[HTML]{FF0000} (-3.19)} &
  40.31 &
  {\color[HTML]{FF0000} (-2.36)} &
  34.87 &
  {\color[HTML]{FF0000} (-3.85)} \\
\multicolumn{1}{l}{} &
  \multicolumn{1}{l|}{} &
  81.22 &
  {\color[HTML]{FF0000} (-2.19)} &
  79.29 &
  {\color[HTML]{FF0000} (-2.66)} &
  24.69 &
  {\color[HTML]{FF0000} (-24.83)} &
  17.29 &
  {\color[HTML]{FF0000} (-28.36)} &
  17.83 &
  {\color[HTML]{FF0000} (-24.84)} &
  10.74 &
  {\color[HTML]{FF0000} (-27.98)} \\ \bottomrule
\end{tabular}
\end{adjustbox}
\end{table}

\begin{table}
\caption{Comparison of $\text{AUBC}_{acc}$(\%) and performance differences (\%p) for different query strategies of standard AL}\label{tab:ablation_standard_al}
\vspace{5pt}
\centering
\footnotesize
\begin{adjustbox}{max width=\textwidth}
\begin{tabular}{@{}ll|cccc|cccc|cccc@{}}
\toprule
\multicolumn{2}{l|}{\textbf{Datasets}} &
  \multicolumn{4}{c|}{\textbf{CIFAR10}} &
  \multicolumn{4}{c|}{\textbf{CIFAR100}} &
  \multicolumn{4}{c}{\textbf{Tiny-ImageNet}} \\ \midrule
\multicolumn{2}{l|}{\textbf{OOD Ratio}} &
  \multicolumn{2}{c|}{\textbf{10\%}} &
  \multicolumn{2}{c|}{\textbf{60\%}} &
  \multicolumn{2}{c|}{\textbf{10\%}} &
  \multicolumn{2}{c|}{\textbf{60\%}} &
  \multicolumn{2}{c|}{\textbf{10\%}} &
  \multicolumn{2}{c}{\textbf{60\%}} \\ \midrule
\multicolumn{2}{l|}{Best $\text{AUBC}_{acc}$ among baselines} &
  82.38 &
  \multicolumn{1}{c|}{-} &
  77.15 &
  - &
  46.20 &
  \multicolumn{1}{c|}{-} &
  34.27 &
  - &
  38.02 &
  \multicolumn{1}{c|}{-} &
  27.28 &
   \\ \midrule
 &
  CONF &
  83.41 &
  \multicolumn{1}{c|}{{\color[HTML]{3C7D22} (+1.03)}} &
  81.95 &
  {\color[HTML]{3C7D22} (+4.80)} &
  49.52 &
  \multicolumn{1}{c|}{{\color[HTML]{3C7D22} (+3.32)}} &
  45.65 &
  {\color[HTML]{3C7D22} (+11.38)} &
  42.67 &
  \multicolumn{1}{c|}{{\color[HTML]{3C7D22} (+4.65)}} &
  38.72 &
  {\color[HTML]{3C7D22} (+11.44)} \\
 &
  LL &
  82.25 &
  \multicolumn{1}{c|}{{\color[HTML]{FF0000} (-0.13)}} &
  80.76 &
  {\color[HTML]{3C7D22} (+3.61)} &
  49.52 &
  \multicolumn{1}{c|}{{\color[HTML]{3C7D22} (+3.32)}} &
  45.77 &
  {\color[HTML]{3C7D22} (+11.50)} &
  40.90 &
  \multicolumn{1}{c|}{{\color[HTML]{3C7D22} (+2.88)}} &
  37.02 &
  {\color[HTML]{3C7D22} (+9.74)} \\
\multirow{-3}{*}{VLPure-AL} &
  CORESET &
  81.69 &
  \multicolumn{1}{c|}{{\color[HTML]{FF0000} (-0.69)}} &
  80.20 &
  {\color[HTML]{3C7D22} (+3.05)} &
  47.48 &
  \multicolumn{1}{c|}{{\color[HTML]{3C7D22} (+1.28)}} &
  44.08 &
  {\color[HTML]{3C7D22} (+9.81)} &
  40.47 &
  \multicolumn{1}{c|}{{\color[HTML]{3C7D22} (+2.45)}} &
  36.81 &
  {\color[HTML]{3C7D22} (+9.53)} \\ \bottomrule
\end{tabular}
\end{adjustbox}
\end{table}

\textbf{Impact of Components.} \cref{tab:ablation_components} presents the ablation study on visual similarity weights and self-temperature tuning. Self-temperature tuning is a method proposed to leverage collected OOD samples effectively. When self-temperature tuning is not applied, the temperature is fixed at 10. The absence of both self-temperature tuning and visual similarity weights leads to a significant performance drop in CIFAR10 and results in poor training outcomes for CIFAR100 and Tiny-ImageNet. This highlights the necessity of applying an optimal temperature. When self-temperature tuning is used to explore the optimal temperature, the method consistently outperforms the baseline across all scenarios. Moreover, even when the optimal temperature is not applied, the inclusion of visual similarity weights achieves better performance than the baseline. This demonstrates that the use of visual similarity weights, derived from labeled data, contributes to improved OOD detection.

\textbf{Standard AL.} VLPure-AL sequentially considers purity first and evaluates informativeness only for samples predicted as ID, enabling the application of various query strategies proposed in standard AL. \cref{tab:ablation_standard_al} shows the performance differences compared to the baseline when applying three standard AL query strategies: CONF, LL, and CORESET. Except for CIFAR10 with an OOD ratio of 10\%, VLPure-AL achieves higher performance than the baseline across all scenarios for all three query strategies.

\begin{figure*}[t!] 
\begin{center}
\includegraphics[width=1.0\linewidth]{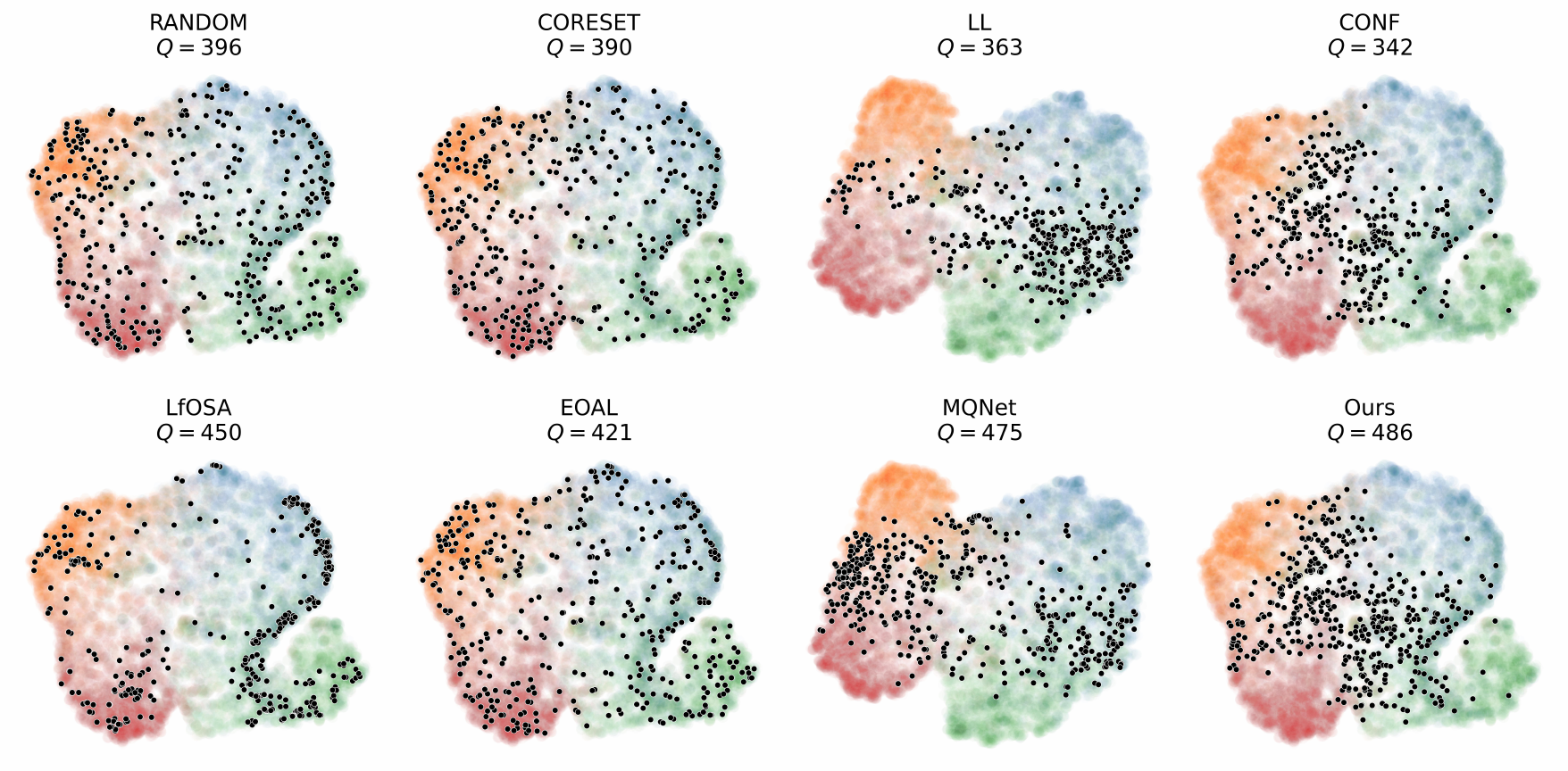}
\end{center}
\vspace{-20pt}
\caption{Visualization for comparing selected queries and informativeness across different query strategies. The trained classifier represents the samples using initial labeled data from CIFAR10, with a 40\% mismatch ratio and 20\% OOD ratio. The black dots represent selected ID data, and the blurred dots represent unlabeled data, with colors indicating class boundaries based on actual labels. LL and MQNet, which use a Loss Prediction Module, show different visual representations from the trained classifiers.}
\label{fig:vis_tsne}
\end{figure*}

\subsection{Visualization of Query Strategy Results} \label{sec:vis}

\cref{fig:vis_tsne} visualizes the ID data and informativeness selected by each query strategy. In the case of RANDOM, the selected data shows no specific pattern and is chosen randomly. In contrast, CORESET selects data with diversity, resulting in less redundancy and more varied data than RANDOM. LL and CONF select data based on uncertainty and choose data near the decision boundaries between classes. However, the query strategies in standard AL result in fewer usable training data due to the selection of OOD data. On the other hand, open-set AL methods select more usable training data than standard AL. However, LfOSA and EOAL, focusing on purity rather than decision boundaries, tend to select less informative data. MQNet selects relatively more ID data than LL, but only in certain decision boundary areas. Conversely, VLPure-AL selects data primarily from high informativeness areas near decision boundaries, similar to CONF, and achieves the highest selection of ID data among the query strategies.

\section{Conclusion}
\label{sec:conclusion}

This study introduces VLPure-AL, a two-stage query strategy designed to minimize cost losses in open-set active learning by improving the accuracy of purity assessments and selecting highly informative data from ID data only. VLPure-AL utilizes a pre-trained CLIPN model to detect OOD data based on linguistic information and visual similarity weights derived from collected ID data. To maximize the accuracy of the pre-trained CLIPN's ID data assessment, VLPure-AL optimizes the temperature parameter through self-temperature tuning based on the labeled ID and OOD data. By accurately selecting ID data from unlabeled data, VLPure-AL can apply a standard AL query strategy to select only highly informative data, reducing cost losses and approximating the conditions assumed in standard AL. 

To validate the efficiency of the proposed method, various open-set datasets were constructed by applying mismatch ratios and OOD ratios to CIFAR10, CIFAR100, and Tiny-ImageNet. Additionally, diverse experiments were conducted using DomainNet by assuming different domains as OOD. Experimental results demonstrate that VLPure-AL consistently achieves the lowest cost losses and highest accuracy across all scenarios. As VLPure-AL conducts purity assessment and data informativeness evaluation sequentially, it can incorporate various query strategies proposed in standard AL after assessing purity. This flexibility indicates that future research can extend the application of VLPure-AL to not only classification but also regression, object detection, and object segmentation in open-set scenarios.


\end{document}